%% file: arxiv.tex
\documentclass[12pt]{article} 
\usepackage{amsmath,amsthm,amssymb,bm,mathrsfs}
\usepackage{graphicx,subfigure}
\usepackage{caption,enumerate,listings,setspace}
\usepackage[OT1]{fontenc}
\usepackage[colorlinks,linkcolor=red,anchorcolor=blue,citecolor=blue]{hyperref}
\usepackage[margin=1in]{geometry}
\usepackage[title]{appendix}
\usepackage{enumitem}
\theoremstyle{plain}

\usepackage[ruled,linesnumbered]{algorithm2e}

\usepackage{booktabs}

\usepackage[numbers]{natbib}

\input{def-smile}

\input{front}

\title{\textbf{
On the Utility Recovery Incapability of Neural Net-based Differential Private Tabular Training Data Synthesizer under Privacy Deregulation}} 

\author
{
Yucong Liu\thanks{Master Student, Department of Statistics, University of Chicago, IL, 60637. Email:yucongliu@uchicago.edu},
Chi-Hua Wang \thanks{Postdoctoral Scholar, Department of Statistics,  UCLA, CA, 90095. Email: chihuawang@ucla.edu},
and
Guang Cheng\thanks{Professor, Department of Statistics, UCLA, CA, 90095. Email:guangcheng@ucla.edu}
}

\date{}
\begin{document} 

\maketitle

\begin{abstract}
Devising procedures for auditing generative model privacy-utility tradeoff is an important yet unresolved problem in practice. Existing works concentrates on investigating the privacy constraint side effect in terms of utility degradation of the \textit{train on synthetic, test on real} paradigm of synthetic data training. We push such understanding on privacy-utility tradeoff to next level by observing the privacy deregulation side effect on synthetic training data utility. Surprisingly, we discover the Utility Recovery Incapability of \texttt{DP-CTGAN} and \texttt{PATE-CTGAN} under privacy deregulation, raising concerns on their practical applications.
The main message is "Privacy Deregulation does NOT always imply Utility Recovery".
\end{abstract}


\section{Introduction}
\label{sec:intro}

\input{z1_Intro.tex}

\section{Relate Work}
\label{sec:relatedWork}
\input{z2_RelatedWork.tex}

\section{Approach and Evaluation Framework}
\label{sec:method}
\input{z3_Method.tex}

\section{Evaluation Results}
\label{sec:result}
\input{z4_Result.tex}

\section{Conclusion}
\label{sec:conclusion}
\input{z5_Conclusion.tex}

\clearpage
\bibliographystyle{unsrt}
\nocite{*}
\bibliography{zr_utility_recovery}

\clearpage
\appendix

\input{z6_Appendix.tex}

\end{document}

%% file: def-smile.tex
\newcommand{\CC}{\mathbb{C}}

\newcommand{\RR}{\mathbb{R}}

\numberwithin{example}{section}



%% file: front.tex
\usepackage[utf8x]{inputenc}
\usepackage{mathrsfs}
\usepackage{amssymb}
\usepackage{amsfonts}
\usepackage{amsmath}
\usepackage{amsthm,verbatim}
\usepackage{wrapfig}
\usepackage{multirow}
\usepackage{longtable}
\usepackage{enumerate}
\usepackage{color, graphicx}
\usepackage{tikz}

\def\boxit#1{\vbox{\hrule\hbox{\vrule\kern6pt
          \vbox{\kern6pt#1\kern6pt}\kern6pt\vrule}\hrule}}



%% file: z1_Intro.tex

Differential private synthetic training data receives rising attentions from both academics and industry \cite{ jordon2022synthetic, mckenna2021winning, cummings2018role}. Such attention is due to the capability of differential private synthetic training data to train machine learning models in a way that aligns modern privacy regulations (e.g. GDPR \cite{EUdataregulations2018} or CCPA). The capability differential private synthetic training data to support trustworthy machine learning lifecycle is fruitful and hence invites both scientists and practitioners to investigate the pros and cons of adopting differential private synthetic training data \cite{jordon2022synthetic, visani2022enabling}. In particular, \textbf{what are the costs and benefits for \textit{machine learning model training} if we replace real training dataset with differential private synthetic training data?}


Indeed, the benefits is a machine learning training process with individual-level privacy protection, and the costs is performance degradation \cite{hittmeir2019utility}. Such observation depreciates the practical usage of differential private synthetic training data, motivating us to investigate a more involved study of the privacy-utility tradeoff \cite{alaa2022faithful, benchmarking} behind synthetic data training. In particular, our investigation focus on the setting of synthesizing differential private tabular training dataset for machine learning classifier training. Given intrinsic individual-level privacy protection by replacing real training data with differential private synthetic training data, this paper is guided by the following question: \textbf{which family of differential private tabular training data synthesizer have the best utility recovery capability?}

\begin{wrapfigure}{r}{10cm}
\centering
\includegraphics[width=\linewidth]{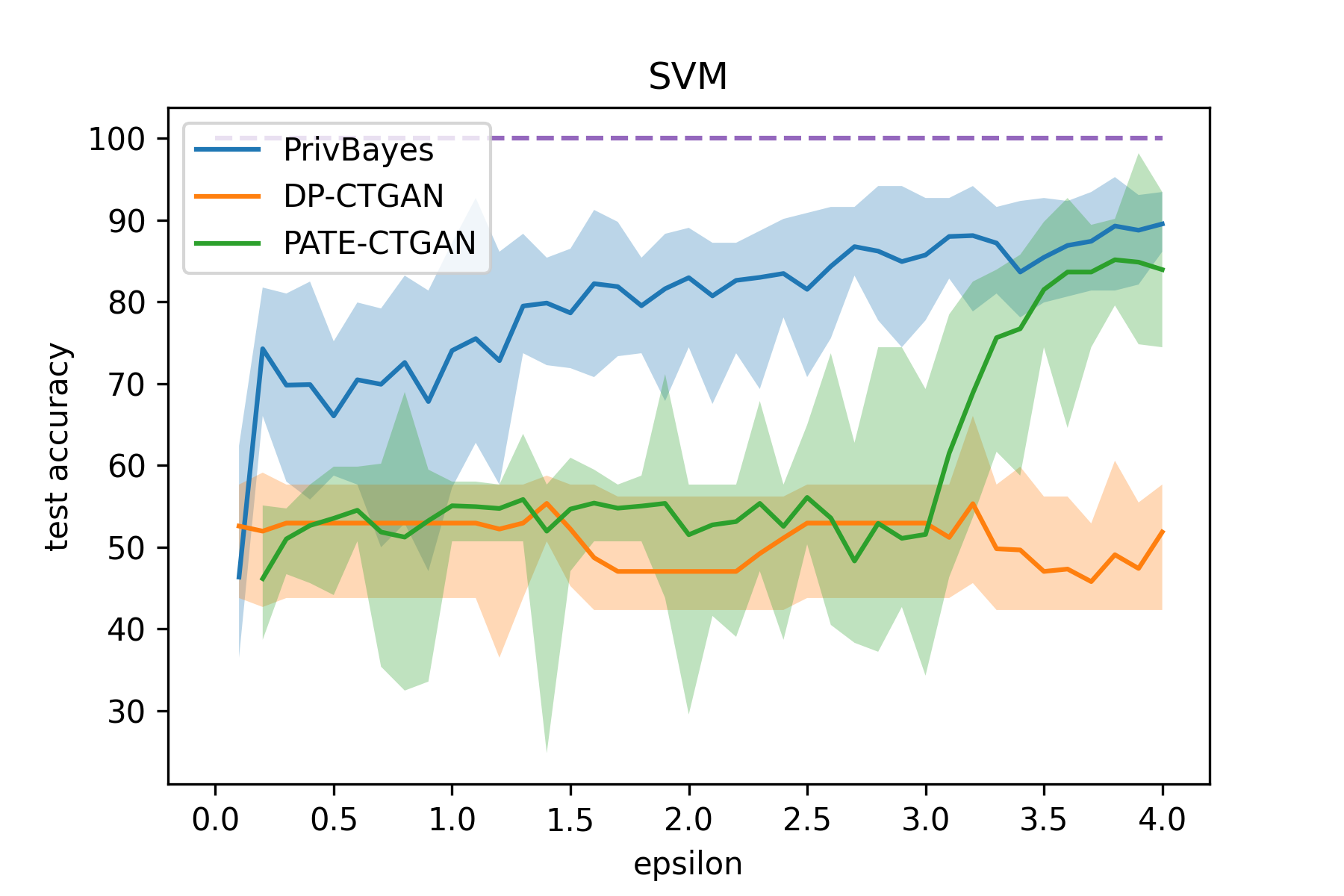}
\caption{Utility Recovery Incapability of \texttt{DP-CTGAN} \cite{dpctgan} and \texttt{PATE-CTGAN} \cite{patectgan} under privacy deregulation when doing synthetic data training for SVM classification model. Fortunately, \texttt{PrivBayes}\cite{PrivBayes} seems immune from such Utility Recovery Incapability under privacy deregulation.}
\label{fig:UtiRecoIncap}
\end{wrapfigure}

At the moment, synthetic data generation research community do not reach an consensus on the mechanism of ensuring differential privacy on synthetic training dataset \cite{visani2022enabling, stadler2022synthetic, abowd2008protective}. In this study, we visit three major family of training data synthetizer with governable differential privacy mechanism: \texttt{DP-GAN} \cite{dpgan}, \texttt{PATE-GAN} \cite{pategan} and \texttt{PrivBayes} \cite{PrivBayes}. These three governable differential privacy mechanism have their counter part on tabular data synthesizers, which are  \texttt{DP-CTGAN} \cite{dpctgan}, \texttt{PATE-CTGAN} \cite{patectgan} and \texttt{DataSynthesizer} \cite{ping2017datasynthesizer}.
Consequently, our aim is to investigate the privacy-utility tradeoff behind these three differential private tabular training data synthesizer.


% ...


In this work, we take the \textit{privacy deregulation} perspective to delineate the privacy-utility tradeoff behind differential private tabular training data synthesizer. Privacy deregulation perspective means that we observe the change of machine learning model utility of synthetic training dataset by changing the privacy budget we required in the differential private training datasets. And we hypothesize that \textit{an trustworthy training data synthesizer should show utility recovery after loosing privacy protection requirement.}
However, Figure \ref{fig:UtiRecoIncap} suggests an upsetting implication that 
\vspace{-2.5mm}\begin{center}
\textbf{the current art of neural net-based differential private tabular training data synthesizer is Utility Recovery Incapable.}
\end{center}\vspace{-2.5mm}
Specifically, we observed that classifier trained on differential private synthetic dataset by \texttt{DP-CTGAN} suffers permanent utility utility damage across different level of privacy budget. On the other hand, \texttt{PATE-CTGAN}-generated differential private training dataset shows certain phrase transition before and after privacy budget $\varepsilon = 3$. 
Such an upsetting observation raises concerns about the usage differential private synthetic training data in wide range tabular data-based machine learning system. Fortunately, bayesian network-based differential private tabular training data synthesizer, \texttt{PrivBayes} \cite{PrivBayes} is observed to be immune from such Utility Recovery Incapability phenomenon.

\textbf{Contribution: Utility Recovery Incapability Universality Disclosure.} In this work, we disclose the Utility Recovery Incapability of neural net-based differential private tabular training data synthesizers. The utility recovery incapability phenomenon raises concerns on whether general utility requirement on differential privacy is a reliable standard to audit the quality of synthetic dataset. Such concern calls for further investigation the specific utility of differential private training data, especially for downstream task-oriented metric.

\textbf{Paper Organization.} We structure this work as follows. 
Section \ref{sec:relatedWork} provides a comprehensive reviews on related research topics across notion of utility and differential privacy (section \ref{subsec:UtilityRef}), differential private tabular data synthesizer (section \ref{subsec:DPtrainingDataRef}) and privacy-utility trade-off (section \ref{subsec:AuditingFrawork}). Section \ref{sec:method} gives experiment details on how to implement differential privacy and privacy deregulation (section \ref{subsec:DPandDereg}), implementation details of synthesizer (section \ref{subsec:DPtrainSynthesis}) and the train on synthetic test on real paradigm (section \ref{subsec:SynTrainTestReal}). Section \ref{sec:result} gives empirical evidence to support our claim that \textbf{the Utility Recovery Incapability phenomenon is universal} across different machine learning models (section \ref{sec:URIdiffModel}), different utility metrics (section \ref{sec:URIdiffMetric}) and different source real dataset (section \ref{sec:URIdiffDataset}). 
Section \ref{sec:conclusion} gives conclusion and discussion on future direction.

%% file: z2_RelatedWork.tex

\subsection{Notion of general utility, specific utility and differential privacy}
\label{subsec:UtilityRef}





\textbf{Notion of General and Specific Utility.} Devising procedures for auditing generative model privacy-utility tradeoff requires more sophisticated thoughts on the notion of utility. In this work, we emphasizes the difference between \textit{general utility}(whether synthetic data distribution is comparable to the original dataset) and \textit{specific utility} (the similarity of downstream task results from the synthetic data and the original data) \cite{snoke2018general}. Also see table 1-1 in \cite{el2020practical} 
for different type of synthetic data and their utility. In this work, we focus on evaluating specific utility of differential private tabular training data synthesizer.





\textbf{Differential Privacy.}
Differential Privacy started with a long line work of algorithmic privacy \cite{dinur2003revealing, dwork2004privacy, blum2005practical}, then 
\cite{dwork2006calibrating} gives the definition of differential privacy. Early work focused on mechanisms of adding noise, like Laplace Mechanism \cite{dwork2006calibrating} and Gaussian Mechanism\cite{dwork2014algorithmic}. These works gave good insight and theoretical support.






\subsection{Differential Private Tabular Training Data Synthesizer}
\label{subsec:DPtrainingDataRef} 


\textbf{Neural Network-based Tabular Data Synthesizer.} Here we review basic and 2 privacy mechanisms to empower neural network-based tabular data synthesizer.
(1) Methods of neural network-based tabular data synthesizer (without differential privacy) include \texttt{CGAN} for tabular data \cite{xu2019modeling}, \texttt{DePart}\cite{mahiou2022dpart}, \texttt{Table-GAN} \cite{tablegan},
\texttt{TGAN}\cite{tgan}
and \texttt{CTAB-GAN}\cite{ctabgan}. (2) On the other hand, those with differential privacy mechenism enable include \texttt{DP-SGD} \cite{DP}, 
\texttt{DPGAN} \cite{dpgan} and extension \cite{frigerio2019differentially},
\texttt{DP-CGAN} \cite{dpcgan}, 
\texttt{GS-WGAN} \cite{gswgan},
differentially private autoencoder-based generative model (\texttt{DP-AuGM}) and differentially private variational autoencoder-based generative model (\texttt{DP-VaeGM})  \cite{chen2018differentially}, \texttt{DP-MERF} \cite{dpmerf}, \texttt{PEARL} \cite{pearl}
\texttt{DP-SYN} \cite{dpsyn}. (3) Under \texttt{PATE} framework \cite{pate}, works include \texttt{PATE-GAN} \cite{pategan},
\texttt{PATE-CTGAN} \cite{patectgan} and \texttt{G-PATE} (Privacy-preserving data Generative model based on the PATE framework) \cite{gpate}.





%










\textbf{Bayesian Network-based Tabular Training Data Synthesizer.} The most representative work of Bayesian Network-based tabular data synthesis methodology is the \texttt{PrivBayes} \cite{PrivBayes}. The \texttt{PrivBayes} utilizes Bayesian network to synthesize (high-dimensional) tabular data with differential privacy guarantee. Read \cite{gogoshin2021synthetic} for detailed probabilistic synthetic data generation framework based on Bayesian Networks.
Bayesian methods serve as a fundamental design to synthesize tabular dataset especially, such as Naive Bayes method for synthesizing healthcare record datasets \cite{avino2018generating} and for generating histogram (\texttt{PrivTree} \cite{PrivTree}).
Works to improve \texttt{PrivBayes} with more sophisticated graphical model design includes \texttt{Private-PGM}(\cite{pgm}), \texttt{APPGM}(\cite{appgm}) and even random markov field-based method \texttt{PrivMRF} (\cite{privmrf}). 
We implement Bayesian network-based synthesizer via the DataSynthesizer \cite{ping2017datasynthesizer}. 









\subsection{Works address Privacy-Utility Tradeoff.}
\label{subsec:AuditingFrawork}





\textbf{Empirical evaluation framework for privacy-utility trade-off.} Here we review existing works on evaluating synthetic datasets privacy-utility trade-off with empirical studies. \cite{dp&pate} compares \texttt{DP-SGD} and \texttt{PATE} with testing accuracy of ML models trained on synthetic dataset. 
\cite{patectgan} adopts AUCROC and F1-score as utility metric to evaluate synthetic ranking agreement over five classification models at different privacy regulation level ($\varepsilon= [0.01, 0.1, 0.5, 1.0, 3.0, 6.0, 9.0]$). 
In additions, \texttt{PATE-CTGAN} \cite{patectgan} and \texttt{DP-GAN} \cite{dpgan} compared Wasserstein distance for different privacy regulation level $\varepsilon = 9.6, 14, 29$. Our work investigates the privacy-utility tradeoff through privacy derregulation perspective, by changing $\varepsilon$ gradually from $0.1$ to $4.0$, to see how fast the utility recover for each family of tabular training datasets synthesizer.

\textbf{Auditing framework for privacy-utility trade-off.} Privacy-Utility trade-off of synthetic datasets is hard to predict \cite{stadler2022synthetic}. 
\cite{alaa2022faithful} evaluates the general utility (fidelity, diversity, generalization) but not into specific utility of generative models. \cite{benchmarking} also evaluate differentially private synthetic tabular data, on both general utility (individual attribute distribution, pairwise correlation) and specific utility (accuracy of ML classification).
Further, \cite{nanayakkara2022visualizing} provides visualization and \cite{lopuhaa2021privacy, zhong2022privacy} gives formal framework to study the privacy-utility tradeoff.  We remarks that \cite{dp&pate} also observed different patterns of privacy-utility tradeoff about \texttt{DP-SGD} and \texttt{PATE} from general utility perspective, while our work reveals the Utility Recovery Incapability from the specific utility perspective. 











%% file: z3_Method.tex
In this section, we give full details on our evaluation framework to disclose the Utility Recovery Incapability phenomenon of neural net-based differential private tabular data training. At section \ref{subsec:DPandDereg}, we formulate the concept of differential privacy and the procedure of privacy deregulation. 
At section \ref{subsec:DPtrainSynthesis}, we gives essential implementation details on synthesizing differential private tabular training datasets. Section \ref{subsec:SynTrainTestReal} establishes the synthetic training and real data testing procedure of our results.

\subsection{Differential Privacy and Privacy Deregulation Synthetic Training}
\label{subsec:DPandDereg}

\textbf{Three differential private mechenisms to synthesize training dataset.} We adopt \texttt{DP-CTGAN}, \texttt{PATE-CTGAN} and \texttt{PrivBayes} to synthesize differential private tabular training data with same privacy budget $\varepsilon$ on several source real datasets. We review the differential privacy mechanism behind these three synthesizers family here.
\texttt{DP-CTGAN} \cite{dpctgan} incorporates differential privacy within the \texttt{CTGAN} model. It  tracks the privacy loss with the privacy accountant framework \cite{PrivacyAccountant}. The utility degradation of \texttt{DP-CTGAN}-based training dataset is due to the difficulties of GAN convergence and privacy loss estimation.  
\texttt{PATE-CTGAN} \cite{patectgan} partition source real dataset into $k$ subdatasets to initialize $k$ conditional generators and then to train $k$ differentially private teacher discriminators.
\texttt{PrivBayes} \cite{PrivBayes} adopt Exponential Mechanism to construct differentially private Bayesian network.

\textbf{Privacy derregulation.} 
By \textit{privacy deregulation synthetic training}, we mean keeping training procedure fixed and changing the training dataset with different privacy budgets.
For each differential private tabular training dataset synthesizer, we generate dataset under privacy budgets $\varepsilon \in \{0.1, 0.2, 0.3, 0.4, \cdots ,4.0\}$.

\subsection{Differential Private Tabular Training Data Synthesis.}
\label{subsec:DPtrainSynthesis}

\textbf{Source Real Datasets.} Our experiments are based on four commonly used machine learning datasets from UCI machine learning repository\footnote{ http://archive.ics.uci.edu/ml} and Kaggle\footnote{https://www.kaggle.com/} (see Table~\ref{datasets}). 
\textit{(i) Banknote.} The Banknote Authentication dataset involves predicting whether a given banknote is authentic given a number of measures taken from a photograph. 
\textit{(ii) Iris.} The Iris Dataset contains four features of 50 samples of three species of Iris. 
\textit{(iii) Social Network Ads.} The Social Network Ads is a categorical dataset to determine whether a user purchased a particular product. 
\textit{(iv)Titanic.} The Titanic dataset tells whether a passenger survived the sinking of the Titanic.

We clean the original dataset by deleting identity-related columns like names. Then, we randomly splits the dateset into training data and test data for 10 times, such that the size of the former is $80\%$ of the original data. For each dataset, we apply a pre-processing procedure with label-encoding and standardization by sklearn.\footnote{https://scikit-learn.org/stable/modules/classes.html\#module-sklearn.preprocessing}

\begin{table}
  \caption{Datasets}
  \label{datasets}
  \centering
  \begin{tabular}{lll}
    \toprule
    Name     & Instances     & Classes \\
    \midrule
    Banknote Authentication & 1372 & 2     \\
    Iris    & 150 & 3     \\
    Social Network Ads  & 400 & 2  \\
    Titanic & 1309 & 2  \\
    \bottomrule
  \end{tabular}
\end{table}
\textbf{Implementation detail of Tabular Training Data Synthesis. } DP-CTGAN \cite{dpctgan} and PATE-CTGAN \cite{patectgan} are implemented with open-source library of SmartNoise\footnote{https://github.com/opendp/smartnoise-sdk}. Datasynthesizer (DS) \cite{ping2017datasynthesizer} is a data generator that takes the structure of the Bayesian network in PrivBayes \cite{PrivBayes}. So, we implement this with correlated attribute mode in  DS.\footnote{https://github.com/DataResponsibly/DataSynthesizer} For each $\varepsilon$, we employ these three methods on the original training dataset respectively and generate 3 synthetic datasets with same size.

\textbf{Implementation detail of Privacy derregulation.} 
For each $\varepsilon$, we generate a synthetic training dataset. So, for each synthetic methods, there are 40 synthetic data with different $\varepsilon$ (except for PATE-CTGAN, which fails to generate with $\varepsilon = 0.1, 0.2$ on Banknote and Titanic).

\subsection{Synthetic Data Training and Real Data Testing for Machine Learning Classifier}
\label{subsec:SynTrainTestReal}

\textbf{Model class of Classifier.}
We approach the classification tasks by applying six different machine learning algorithms, namely Na\"ive Bayes, Support Vector Machines (SVM), K-Nearest Neighbours (KNN), Decision Tree, Random Forests and Logistic Regression. All classifiers are implemented with sklearn package\footnote{https://scikit-learn.org/stable/index.html}, and we use standard parameters.

\textbf{Synthetic Data Training procedure} We train each of our 6 classifiers with original training data and synthetic training data respectively. We test these classifiers on the original test data for 6 metrics. 

\textbf{Real Data Testing metric}
We use 6 different metrics in our experiments, including Accuracy, Precision, Recall, F1 Score, Area Under the Receiver Operating Characteristic Curve (AUROC) and Average Precision. All metrics are implemented with sklearn package. We calculate the mean, maximum and minimum of each metric for totally 10 data splits. Because Iris dataset has multi-class, we don't calculate Average Precision for this dataset.

%% file: z4_Result.tex
In this section, we provide empirical evidences to support our claim that 
\vspace{-3mm}\begin{center}
\textbf{the Utility Recovery Incapability phenomenon is universal}. 
\end{center}\vspace{-3mm}
Our evaluation across three major dimension in synthetic data training: Model (Classifier), Metric (Utility) and Data (Source Dataset). Section \ref{sec:URIdiffModel} gives the result across different classifier with Accuracy metric on Banknote source dataset. Section \ref{sec:URIdiffMetric} gives the result across different utility metric with KNN model on Banknote source dataset. Section \ref{sec:URIdiffDataset} gives the result across different source dataset with Random Forest model with Accuracy metric. See Appendix \ref{appendix:model}, \ref{appendix:metric} , \ref{appendix:dataset}  for our extensive investigation results to support the universality of Utility Recovery Incapability phenomenon.

\subsection{Utility Recovery Incapability phenomenon across different Machine Learning Models} 
\label{sec:URIdiffModel}

\begin{figure}[t]
    \centering
    \makebox[\textwidth][c]{\includegraphics[width=1.2\linewidth]{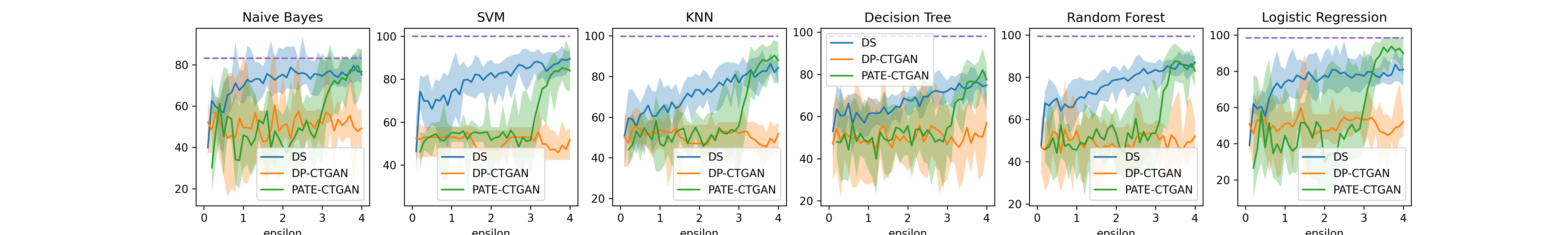}}
    \caption{Utility Recovery Incapability across different Machine Learning Models (Section \ref{sec:URIdiffModel}). Testing accuracy result on Banknote dataset. \texttt{DP-CTGAN} shows utility recovery incapability. \texttt{PATE-CTGAN} shows utility recovery incapability under strict privacy requirement $(\epsilon<3)$ and recover utility under loose privacy requirement $(\epsilon>3)$. \texttt{DataSynthesizer} shows stable utility recovery. Thus, we recommend Bayesian Network-based synthesizer.}
    \label{banknote accuracy}
\end{figure}

Figure \ref{banknote accuracy} shows that Accuracy on Banknote dataset has similar trend across 6 different classifiers. 

First, accuracy of DP-CTGAN is always around $50\%$ no matter which classifier is used. This implies that the classifier with DP-CTGAN synthesized data is just a random guess. Moreover, when $\varepsilon$ increases, Accuracy stays. Thus, DP-CTGAN is incapable for utility recovery.

Second, the test accuracy with PATE-CTGAN starts to recover when $\varepsilon > 3$ for all classifier. And the interval range between maximum and minimum tends to shrink, which implies the test accuracy with PATE-CTGAN becomes stable when $\varepsilon$ increases. We can interpret this as robustness under synthesizing. However, when $\varepsilon < 3$, PATE-CTGAN also fails on the classification and has similar performance as DP-CTGAN. There is no phenomenon of recovery for small $\varepsilon$.

Finally, DataSynthesizer show best performance out of these three synthesizing methods. It has best accuracy, smallest extremum range and also a utility recovery phenomenon. This Bayesian Network method also achieves a acceptable performance when $\varepsilon$ is small.
For more results, see Appendix \ref{appendix:model}.

\subsection{Utility Recovery Incapability phenomenon  across different classification utility metrics}
 \label{sec:URIdiffMetric}

\begin{figure}[t]
    \centering
    \makebox[\textwidth][c]{\includegraphics[width=1.2\linewidth]{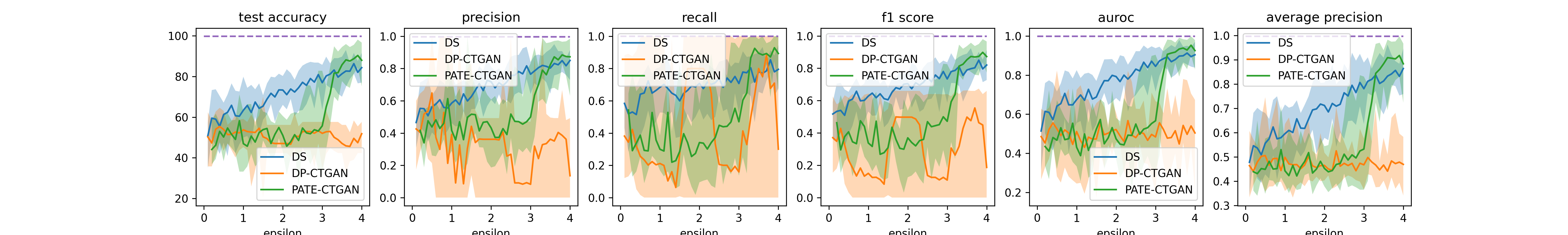}}
    \caption{Utility Recovery Incapability phenomenon across different Classification Utility Metrics (Section \ref{sec:URIdiffMetric}): Results of KNN classifier on Banknote dataset. \texttt{DP-CTGAN} shows different recovery patterns under different utility metrics. \texttt{PATE-CTGAN} shows utility recovery incapability under strict privacy requirement $(\epsilon<3)$ and recover utility under loose privacy requirement $(\epsilon>3)$.\texttt{DataSynthesizer} shows stable utility recovery. Thus, we recommend Bayesian Network-based synthesizer.}
    \label{banknote knn}
\end{figure}

Figure \ref{banknote knn} shows different metrics results on Banknote dataset with KNN classifier. The trends on different metrics are also close.

First, DP-CTGAN stays at a low mean value for all metrics at almost all $\varepsilon$. DP-CTGAN also has a very large range for Precision, Recall and F1 Score. These results imply the prediction with DP-CTGAN is not reliable.  

Second, PATE-CTGAN has consistent performance for each metrics. Only when $\varepsilon > 3$, metrics with PATE-CTGAN rises fast. It still works badly with small $\varepsilon$.

Finally, DataSynthesizer keeps to be the best one and rises steadily with increasing $\varepsilon$. It also has the smallest value range for each $\varepsilon$.
For some classifier, DataSynthesizer has different performance on Recall. To be detailed, some Recall curve doesn't rise but still maintains at a high value level.
For more results, see Appendix \ref{appendix:metric}. 

\subsection{Utility Recovery Incapability phenomenon  across different source real dataset}
\label{sec:URIdiffDataset}

\begin{figure}[t]
    \centering
    \includegraphics[width=1\linewidth]{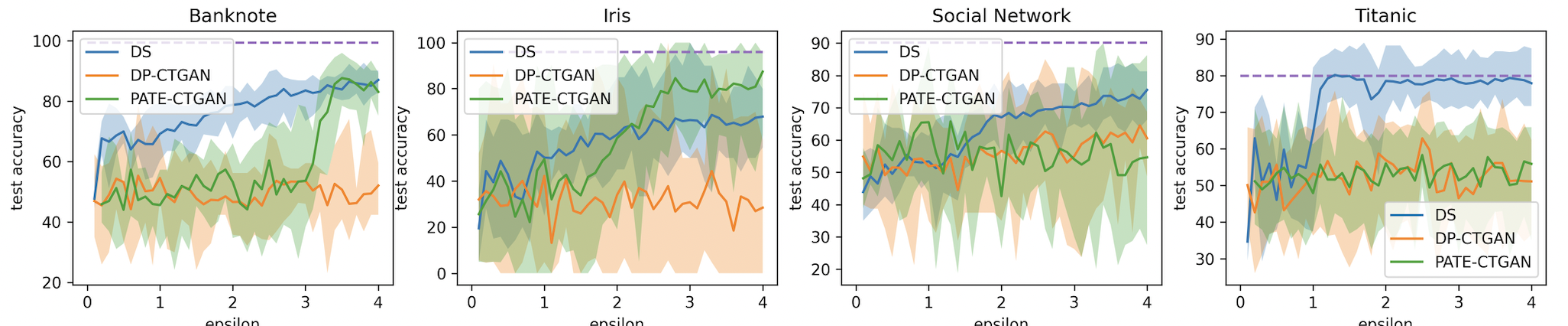}
    \caption{Utility Recovery Incapability phenomenon across different Source Real Dataset (Section \ref{sec:URIdiffDataset}): 
    Testing accuracy results of  Random Forest classifier.
    \texttt{DP-CTGAN} shows utility recovery incapability over 4 source dataset.  \texttt{PATE-CTGAN} shows utility recovery only on Banknote and Iris source dataset under loose privacy requirement. \texttt{DataSynthesizer} shows stable utility recovery over 4 source datasets. Thus, we recommend Bayesian Network-based synthesizer.}
    \label{arf}
\end{figure}

Figure \ref{arf} shows Accuracy with Random Forest on different datasets. Different from previous evaluation, results among different datasets has different trend, especially for PATE-CTGAN.

DP-CTGAN and DataSynthesizer are roughly consistent. DP-CTGAN fails and DataSynthesizer shows good utility recovery phenomenon for all datasets.

Differently, PATE-CTGAN works well on Banknote and Iris but fails on Social Network Ads and Titanic. Moreover, the accuracy with PATE-CTGAN starts to rise at different $\varepsilon$, about 3 on Banknote and about 1.5 on Iris. When trained with PATE-CTGAN, accuracy on Banknote also has much smaller value  range than accuracy on Iris. Therefore, the performance of PATE-CTGAN is strongly affected by the dataset.
For more results, see Appendix \ref{appendix:dataset}.

%% file: z5_Conclusion.tex
In this work, we provide a practical evaluation of differential private tabular data synthesizer for synthetic training of machine learning classifier. Our evaluation covers data, models, and metrics to support the claim that the utility recovery incapability phenomenon is universal. 

One promising future work direction is to devise \textit{utility recovery monitoring process} for differential private tabular data synthesizer. Such monitoring process is indispensable for synthetic data practitioner, and we believe more studies on this regard would empower machine learning research toward trustworthy usage of synthetic data methodology in near future. 



%% file: z6_Appendix.tex
\section{More Results for Utility Recovery Incapability phenomenon across different Machine Learning Models}\label{appendix:model}
\begin{figure}[ht]
    \centering
    \makebox[\textwidth][c]{\includegraphics[width=1.25\linewidth]{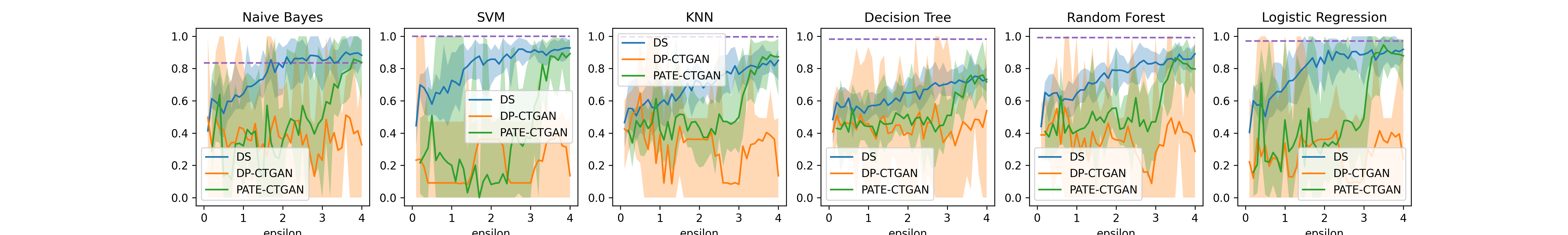}}
    \caption{Banknote Precision}
    \label{banknote precision}
\end{figure}

\begin{figure}[ht]
    \centering
    \makebox[\textwidth][c]{\includegraphics[width=1.25\linewidth]{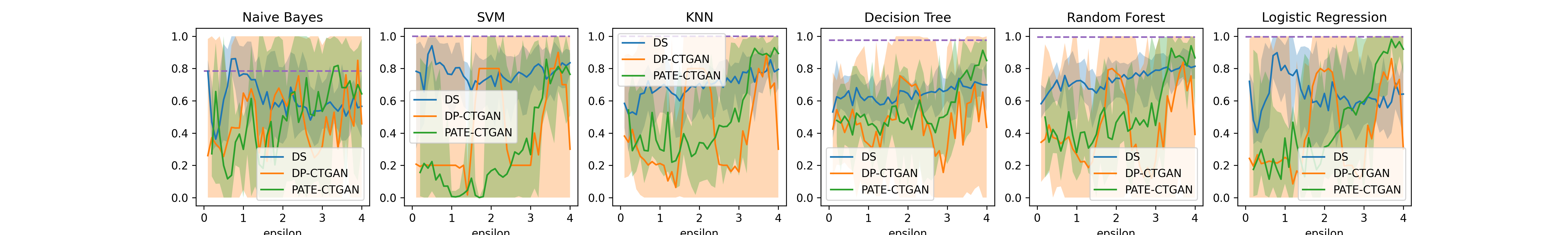}}
    \caption{Banknote Recall}
    \label{banknote recall}
\end{figure}

\begin{figure}[ht]
    \centering
    \makebox[\textwidth][c]{\includegraphics[width=1.25\linewidth]{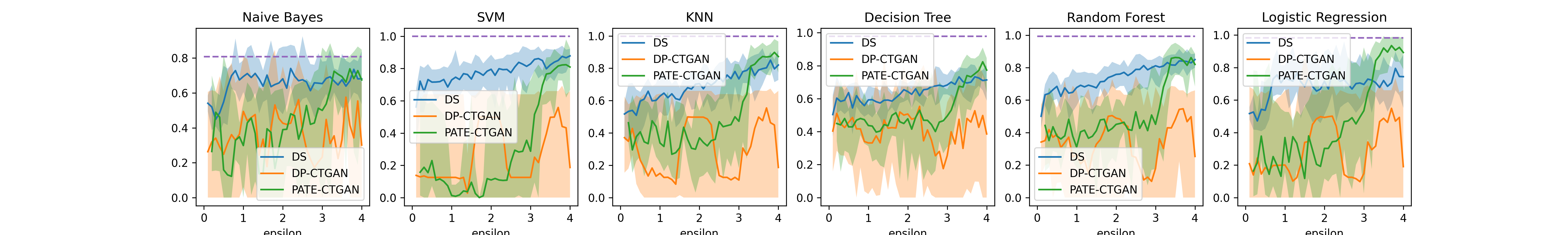}}
    \caption{Banknote F1 Score}
    \label{banknote f1}
\end{figure}

\begin{figure}[ht]
    \centering
    \makebox[\textwidth][c]{\includegraphics[width=1.25\linewidth]{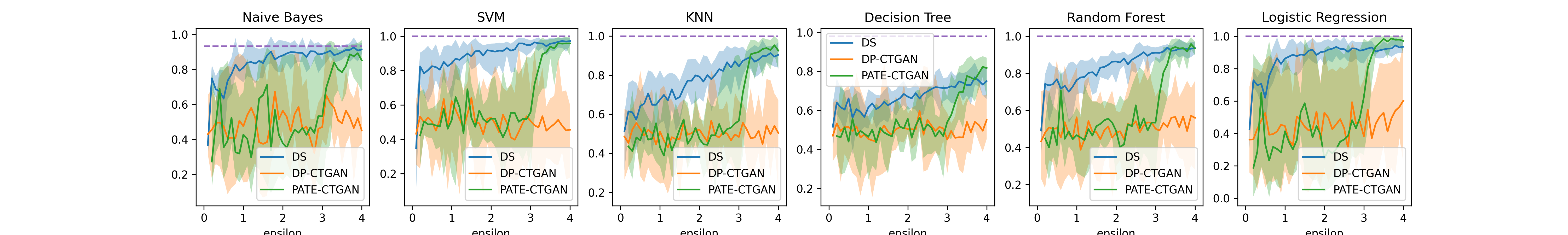}}
    \caption{Banknote AUROC}
    \label{banknote auroc}
\end{figure}

\begin{figure}[ht]
    \centering
    \makebox[\textwidth][c]{\includegraphics[width=1.25\linewidth]{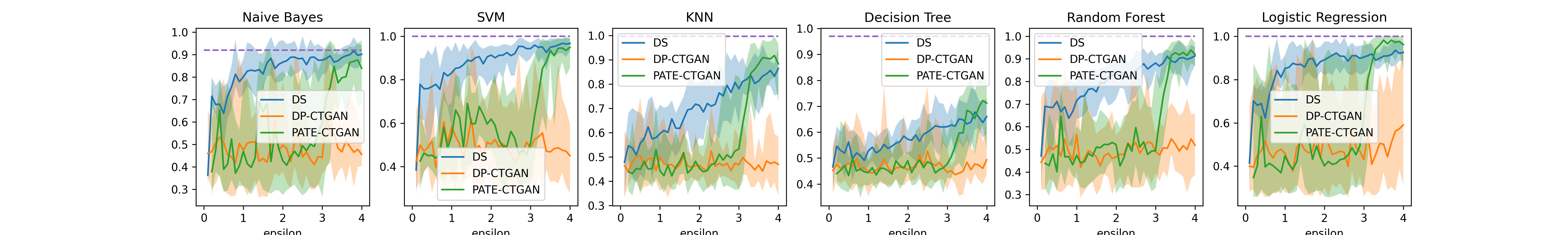}}
    \caption{Banknote Average Precision}
    \label{banknote ap}
\end{figure}

\textbf{Banknote Authentication}
Instead of Accuracy \ref{banknote accuracy} shown before, we here present results for other four metrics. Precision \ref{banknote precision}, F1 Score \ref{banknote f1}, AUROC \ref{banknote auroc} and Average Precision \ref{banknote ap} show similar phenomenon as shown in Accuracy \ref{banknote accuracy}. Both DataSynthesizer and PATE-CTGAN have the trend of utility recovery. And DataSynthesizer has the best performance especially when $\varepsilon$ is small. As mentioned before, Recall \ref{banknote recall} has slightly different result for DataSynthesizer but it still maintains at a good level.

\begin{figure}[ht]
    \centering
    \makebox[\textwidth][c]{\includegraphics[width=1.25\linewidth]{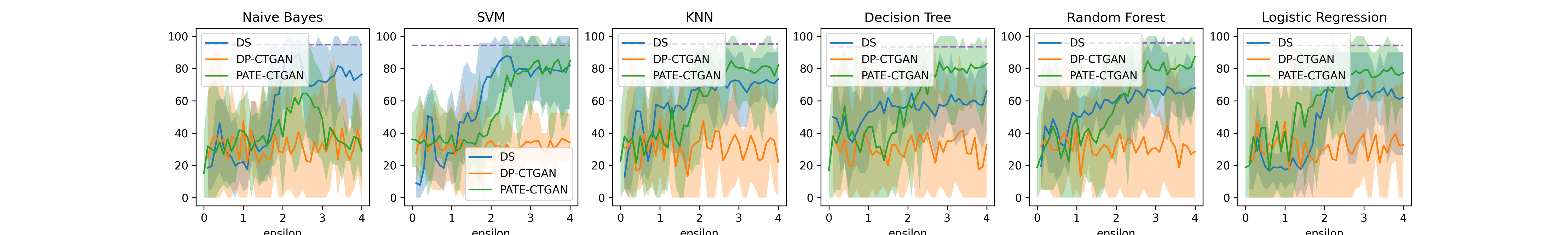}}
    \caption{Iris Accuracy}
    \label{iris accu}
\end{figure}

\begin{figure}[ht]
    \centering
    \makebox[\textwidth][c]{\includegraphics[width=1.25\linewidth]{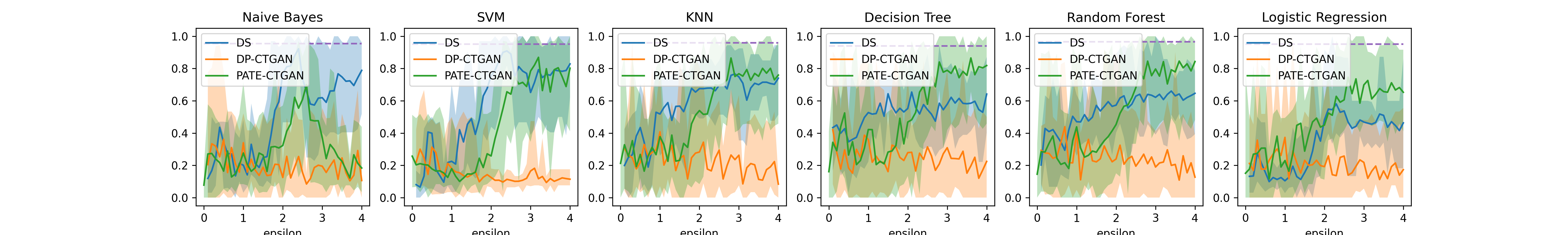}}
    \caption{Iris Precision}
    \label{iris precision}
\end{figure}

\begin{figure}[ht]
    \centering
    \makebox[\textwidth][c]{\includegraphics[width=1.25\linewidth]{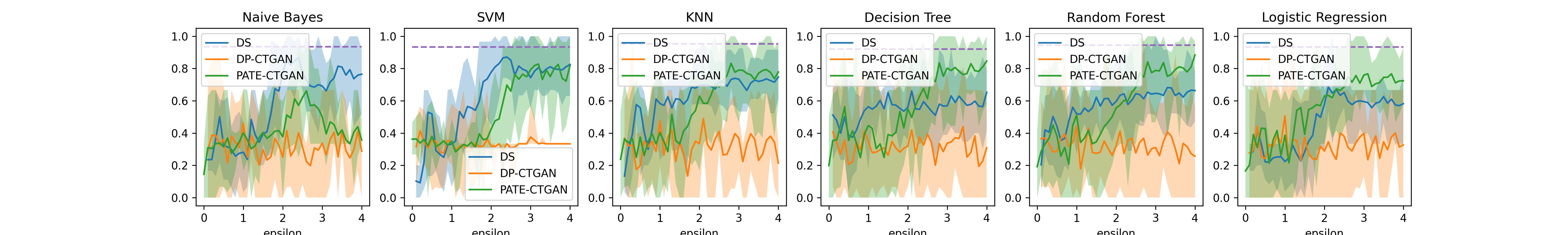}}
    \caption{Iris Recall}
    \label{iris recall}
\end{figure}

\begin{figure}[ht]
    \centering
    \makebox[\textwidth][c]{\includegraphics[width=1.25\linewidth]{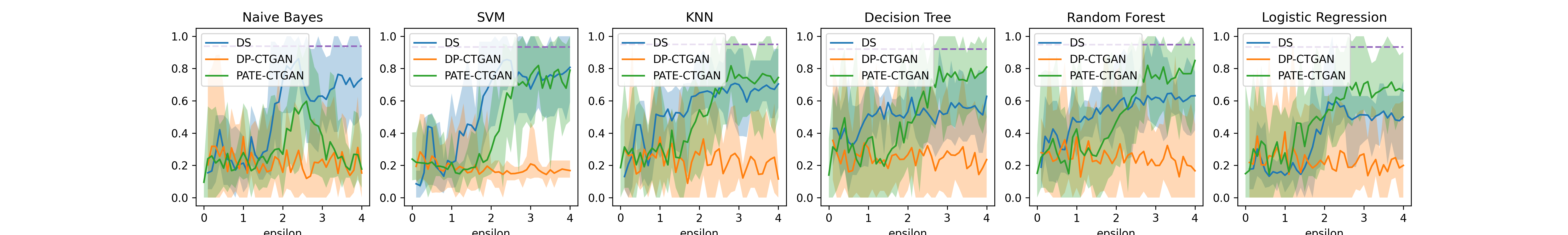}}
    \caption{Iris F1 Score}
    \label{iris f1}
\end{figure}

\begin{figure}[ht]
    \centering
    \makebox[\textwidth][c]{\includegraphics[width=1.25\linewidth]{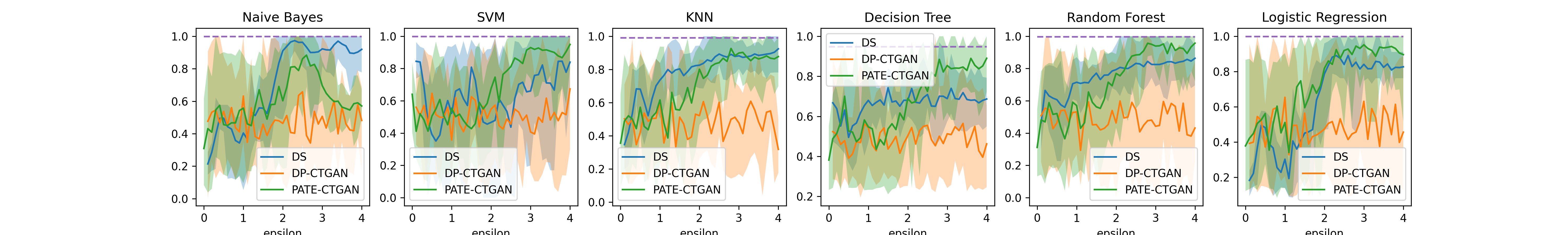}}
    \caption{Iris AUROC}
    \label{iris auroc}
\end{figure}

\textbf{Iris}
Iris is a three-classes dataset. So, there is no average precision on Iris.  We only test other five metrics on this Iris dataset, shown in figure \ref{iris accu}, \ref{iris precision}, \ref{iris recall}, \ref{iris f1} and \ref{iris auroc}. For other metrics except for Accuracy, we calculate unweighted mean for each label as the metric.
Results on Iris is slightly different from previous result. Both DataSynthesizer and PATE-CTGAN shows the phenomenon of utility recovery and PATE-CTGAN exceed DataSynthesizer when $\epsilon$ increases.  Same as results on Banknote dataset, only PATE-CTGAN has the recovery on Recall. There are also difference between each classifier. DataSynthesizer works better with Na\"ive Bayes and SVM on Accuracy, Precision, Recall and F1 Score. PATE-CTGAN seems to fail with Na\"ive Bayes. This implies Bayesian Network based methods may work better with Bayesian methods.

\begin{figure}[ht]
    \centering
    \makebox[\textwidth][c]{\includegraphics[width=1.25\linewidth]{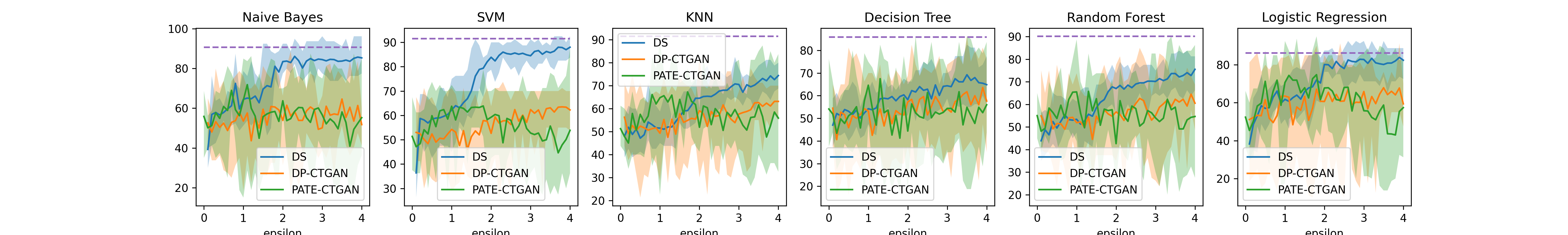}}
    \caption{Social Network Accuracy}
    \label{sn accu}
\end{figure}

\begin{figure}[ht]
    \centering
    \makebox[\textwidth][c]{\includegraphics[width=1.25\linewidth]{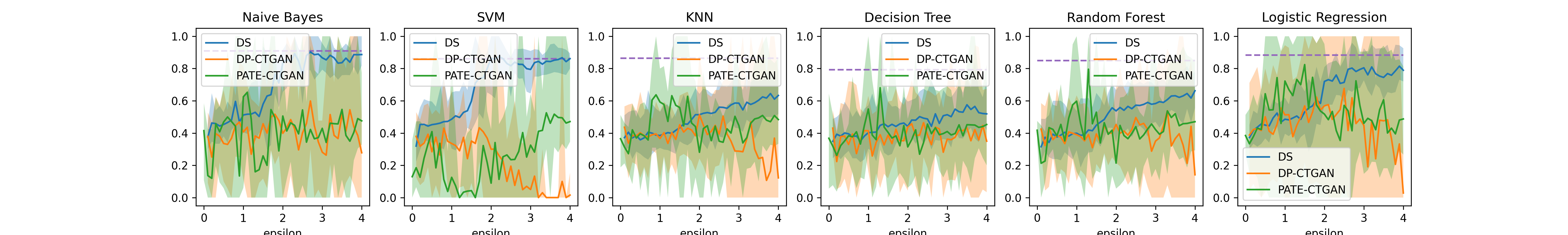}}
    \caption{Social Network Precision}
    \label{sn precision}
\end{figure}

\begin{figure}[ht]
    \centering
    \makebox[\textwidth][c]{\includegraphics[width=1.25\linewidth]{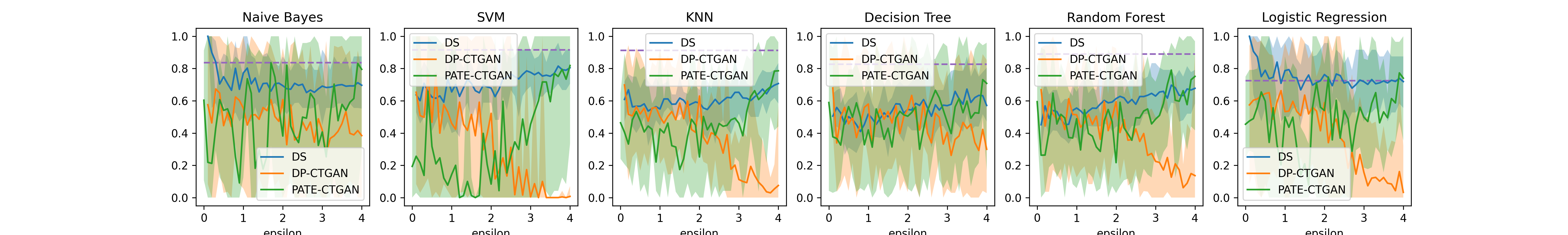}}
    \caption{Social Network Recall}
    \label{sn recall}
\end{figure}

\begin{figure}[ht]
    \centering
    \makebox[\textwidth][c]{\includegraphics[width=1.25\linewidth]{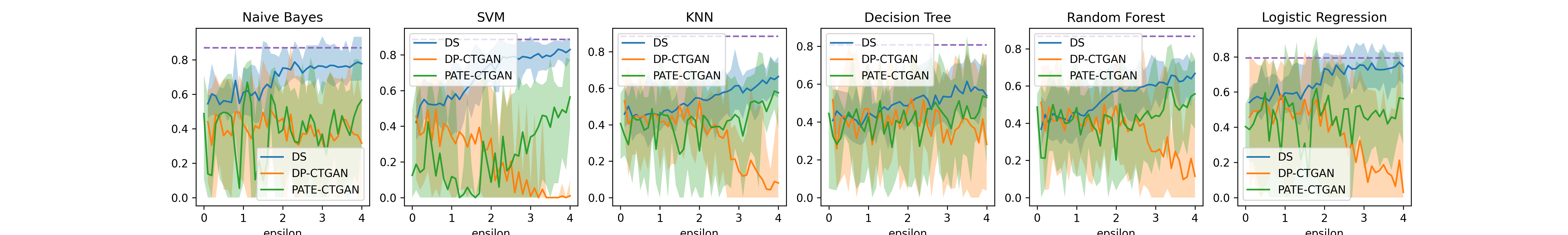}}
    \caption{Social Network F1 Score}
    \label{sn f1}
\end{figure}

\begin{figure}[ht]
    \centering
    \makebox[\textwidth][c]{\includegraphics[width=1.25\linewidth]{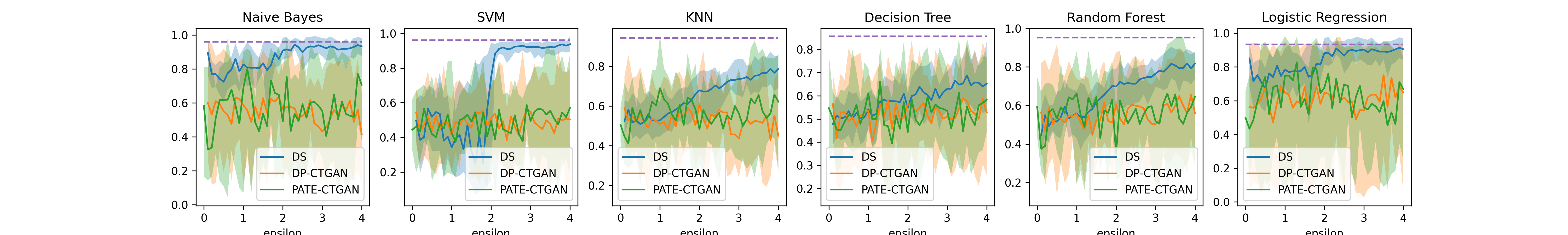}}
    \caption{Social Network AUROC}
    \label{sn auroc}
\end{figure}

\begin{figure}[ht]
    \centering
    \makebox[\textwidth][c]{\includegraphics[width=1.25\linewidth]{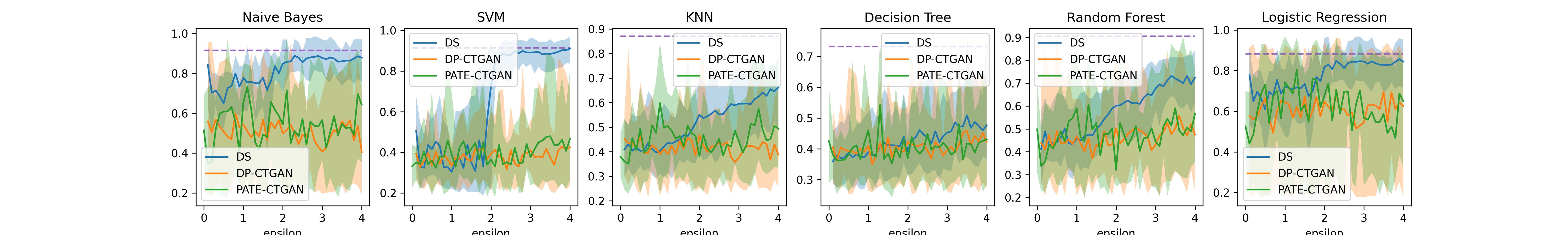}}
    \caption{Social Network Average Precision}
    \label{sn ap}
\end{figure}

\textbf{Social Network Ads}
On this dataset, results are quite different from results on previous datasets. PATE-CTGAN fails on this dataset, as well as DP-CTGAN. Both of them only have accuracy around 0.5, no utility recovery for all metrics and a very large range between minimum and maximum. To the opposite, DataSynthesizer achieve best performance, which is very close to result on original training data. It works better with Na\"ive Bayes, SVM and Logistic Regression, compared with other classifier. When $\epsilon$ increases, performance recover except for Recall. However, this method has Recall more than 0.6 for most $\epsilon$ with Na\"ive Bayes, SVM and Logistic Regression. DataSynthesizer also has a small range between minimum and maximum. As a result, it is accurate and robust on Social Network Ads. See Figures below for Accuracy \ref{sn accu}, Precision \ref{sn precision}, Recall \ref{sn recall}, F1 Score \ref{sn f1}, AUROC \ref{sn auroc} and Average Precision \ref{sn ap}. 

\begin{figure}[ht]
    \centering
    \makebox[\textwidth][c]{\includegraphics[width=1.25\linewidth]{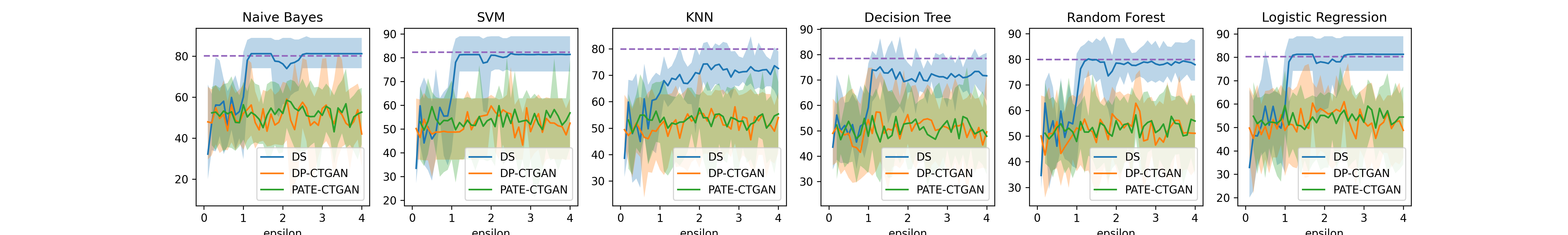}}
    \caption{Titanic Accuracy}
    \label{t accu}
\end{figure}

\begin{figure}[ht]
    \centering
    \makebox[\textwidth][c]{\includegraphics[width=1.25\linewidth]{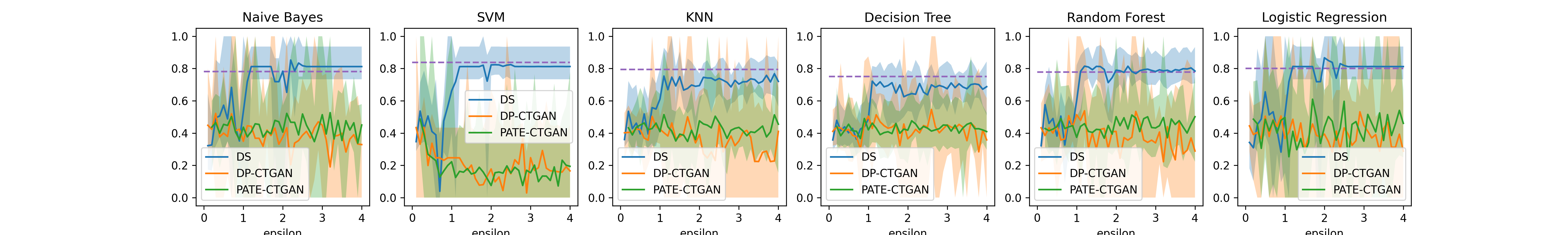}}
    \caption{Titanic Precision}
    \label{t precision}
\end{figure}

\begin{figure}[ht]
    \centering
    \makebox[\textwidth][c]{\includegraphics[width=1.25\linewidth]{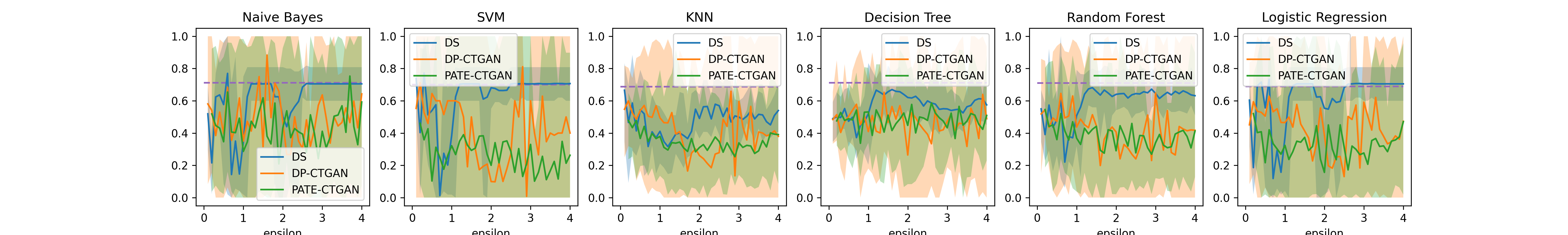}}
    \caption{Titanic Recall}
    \label{t recall}
\end{figure}

\begin{figure}[ht]
    \centering
    \makebox[\textwidth][c]{\includegraphics[width=1.25\linewidth]{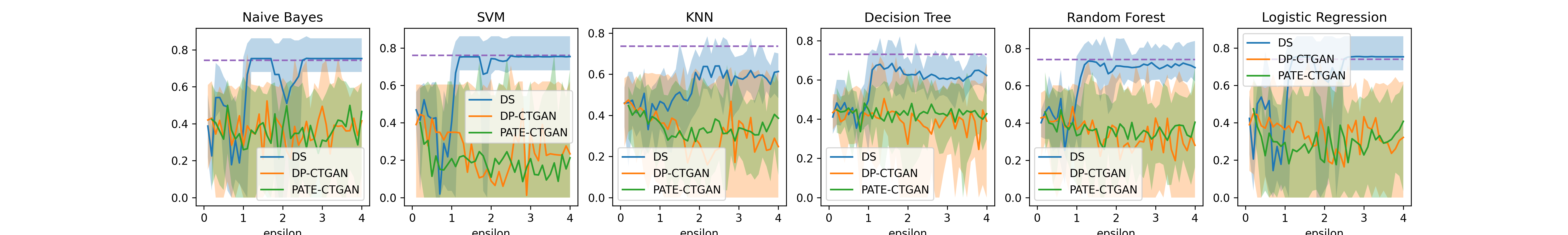}}
    \caption{Titanic F1 Score}
    \label{t f1}
\end{figure}

\begin{figure}[ht]
    \centering
    \makebox[\textwidth][c]{\includegraphics[width=1.25\linewidth]{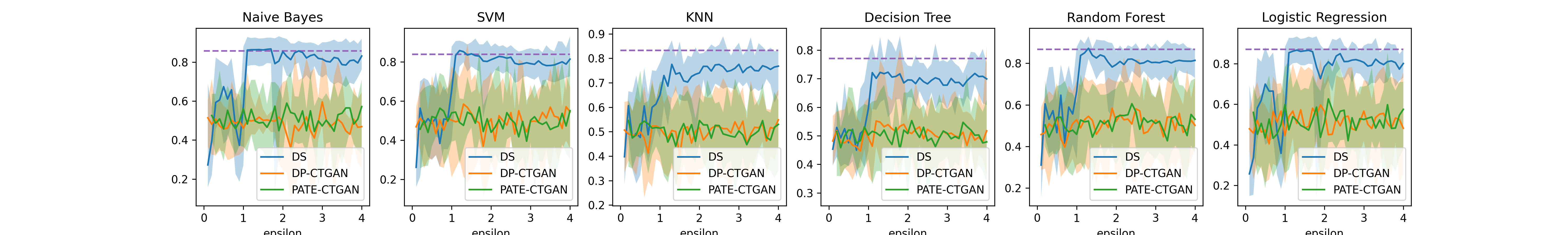}}
    \caption{Titanic AUROC}
    \label{t auroc}
\end{figure}

\begin{figure}[ht]
    \centering
    \makebox[\textwidth][c]{\includegraphics[width=1.25\linewidth]{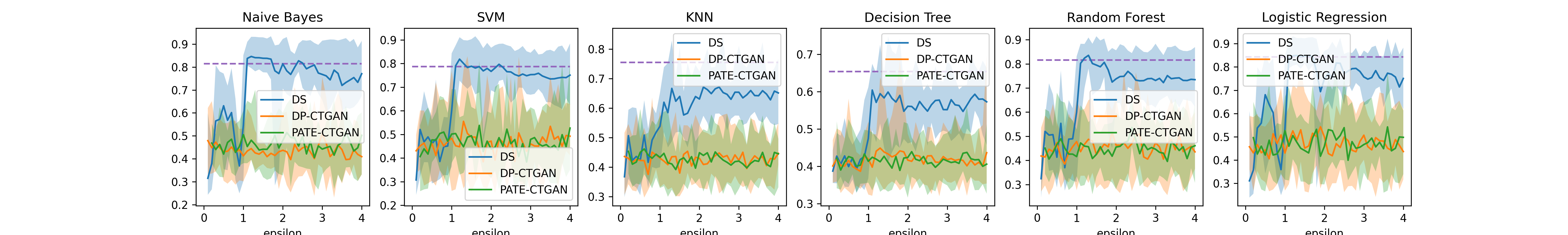}}
    \caption{Titanic Average Precision}
    \label{t ap}
\end{figure}

\textbf{Titanic}
Results on Titanic is very similar with results on Social Network Ads. Only DataSynthesizer has good performance. Interestingly, when $\varepsilon = 1.0$, the performance of DataSynthesizer is already very close to the performance of real training data. See figures for Accuracy \ref{t accu}, Precision \ref{t precision}, Recall \ref{t recall}, F1 Score \ref{t f1}, AUROC \ref{t auroc} and Average Precision \ref{t ap} on Titanic dataset.

\clearpage

\section{More Results for Utility Recovery Incapability phenomenon  across different classification utility metrics}\label{appendix:metric}
\begin{figure}[ht]
    \centering
    \makebox[\textwidth][c]{\includegraphics[width=1.25\linewidth]{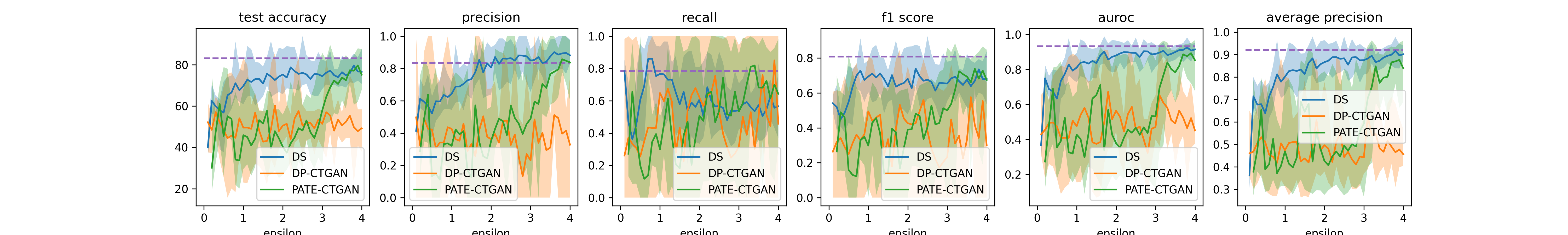}}
    \caption{Banknote with Na\"ive Bayes}
    \label{banknote nb}
\end{figure}

\begin{figure}[ht]
    \centering
    \makebox[\textwidth][c]{\includegraphics[width=\linewidth]{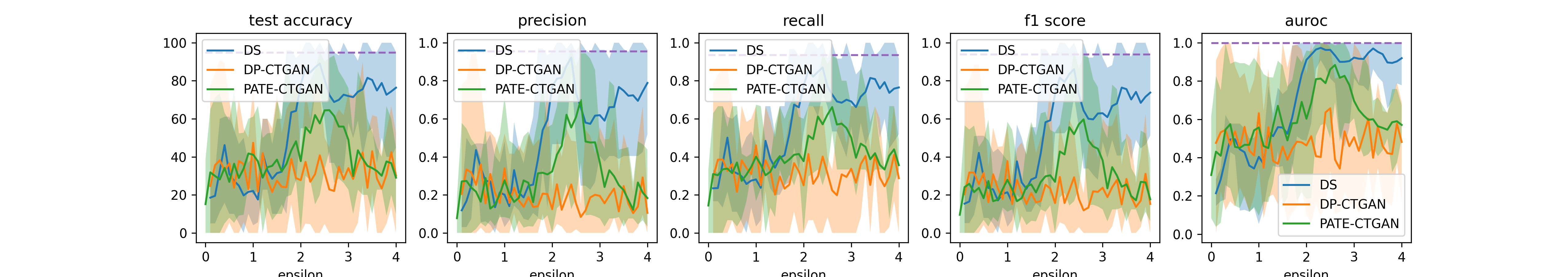}}
    \caption{Iris with Na\"ive Bayes}
    \label{iris nb}
\end{figure}

\begin{figure}[ht]
    \centering
    \makebox[\textwidth][c]{\includegraphics[width=1.25\linewidth]{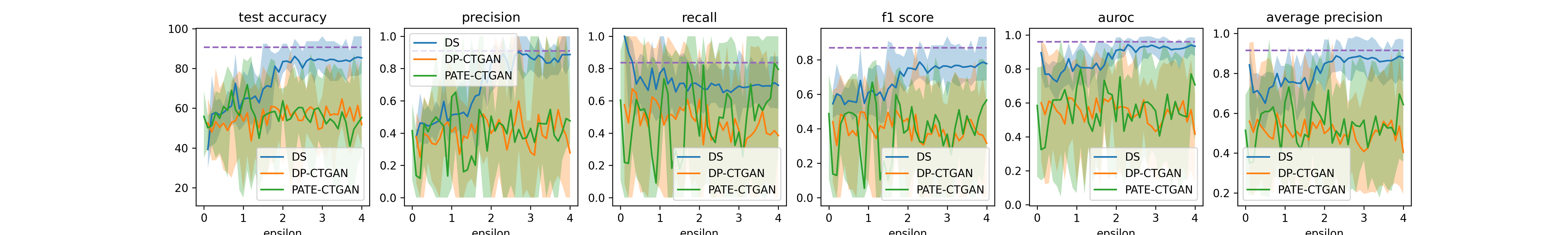}}
    \caption{Social Network with Na\"ive Bayes}
    \label{sn nb}
\end{figure}

\begin{figure}[ht]
    \centering
    \makebox[\textwidth][c]{\includegraphics[width=1.25\linewidth]{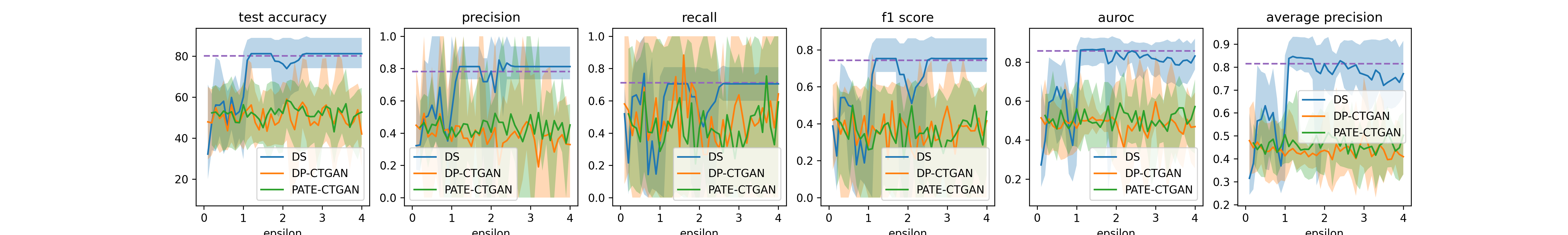}}
    \caption{Titanic with Na\"ive Bayes}
    \label{t nb}
\end{figure}

\textbf{Na\"ive Bayes}
Figures show the performance of Na\'ive Bayes on Banknote \ref{banknote nb}, Iris \ref{iris nb}, Social Network Ads \ref{sn nb} and Titanic \ref{t nb}. DataSynthesizer gets very close to the results on original training data, which are much better  than other two GAN-based methods,

\begin{figure}[ht]
    \centering
    \makebox[\textwidth][c]{\includegraphics[width=1.25\linewidth]{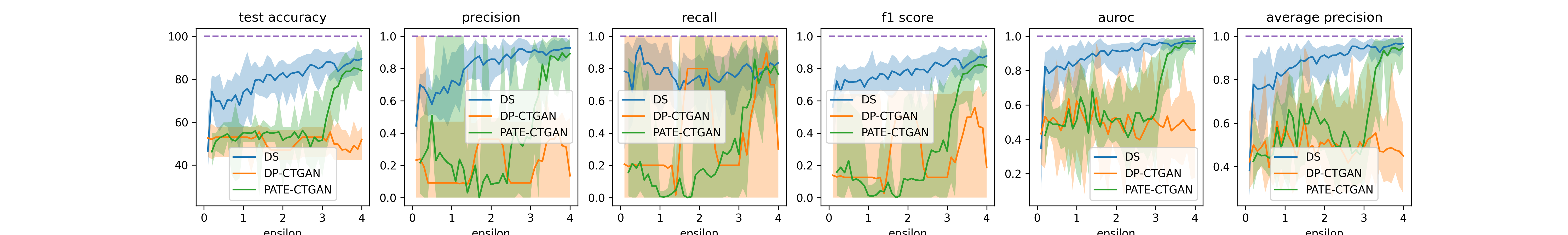}}
    \caption{Banknote with SVM}
    \label{banknote svm}
\end{figure}

\begin{figure}[ht]
    \centering
    \makebox[\textwidth][c]{\includegraphics[width=\linewidth]{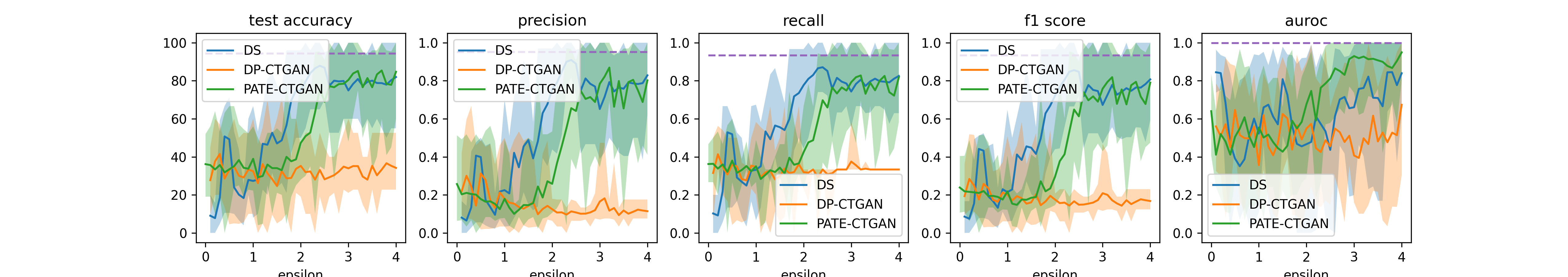}}
    \caption{Iris with SVM}
    \label{iris svm}
\end{figure}

\begin{figure}[ht]
    \centering
    \makebox[\textwidth][c]{\includegraphics[width=1.25\linewidth]{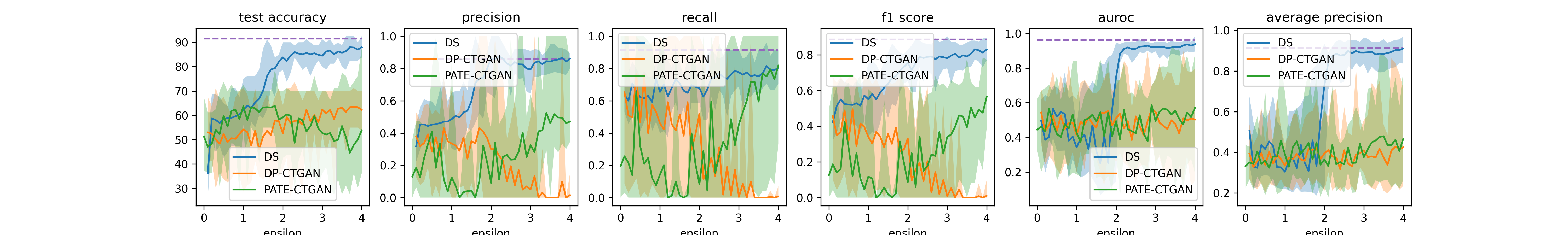}}
    \caption{Social Network with SVM}
    \label{sn svm}
\end{figure}

\begin{figure}[ht]
    \centering
    \makebox[\textwidth][c]{\includegraphics[width=1.25\linewidth]{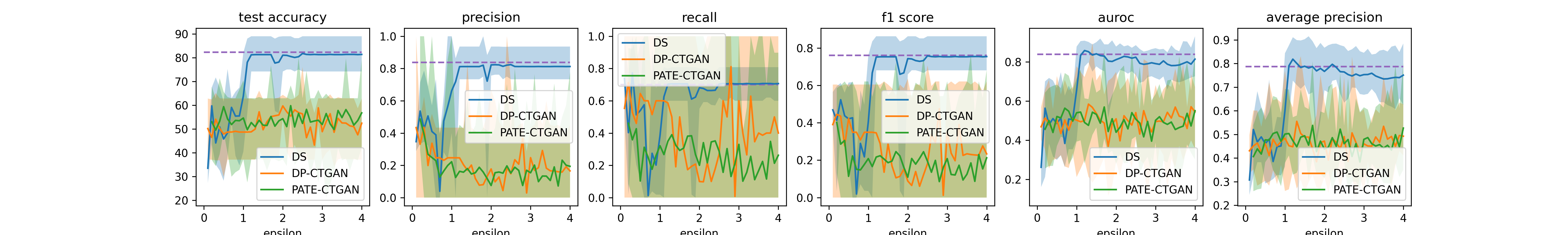}}
    \caption{Titanic with SVM}
    \label{t svm}
\end{figure}

\textbf{SVM}
SVM has similar results with Na\"ive Bayes. DataSynthesizer is the best one. PATE-CTGAN can only get close to DataSynthesizer with large $\varepsilon$ on Banknote \ref{banknote svm} and Iris \ref{iris svm} and become as bad as DP-CTGAN on Social Network \ref{sn svm} and Titanic \ref{t svm}.

\begin{figure}[ht]
    \centering
    \makebox[\textwidth][c]{\includegraphics[width=\linewidth]{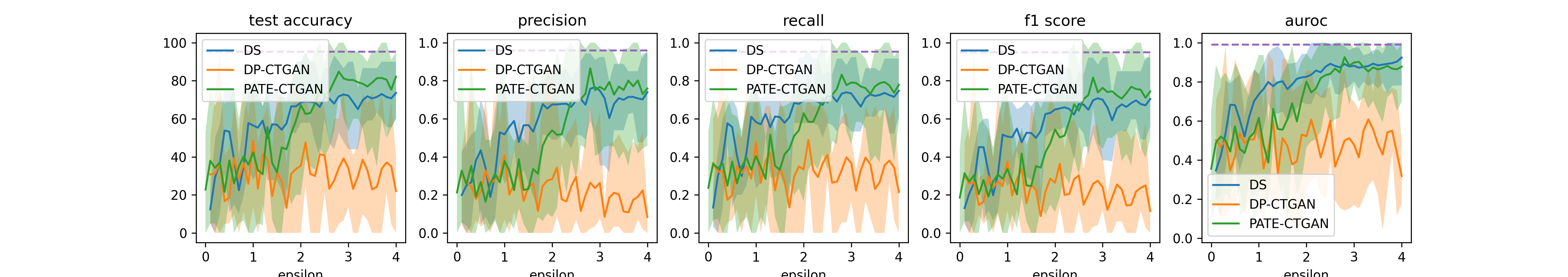}}
    \caption{Iris with KNN}
    \label{iris knn}
\end{figure}

\begin{figure}[ht]
    \centering
    \makebox[\textwidth][c]{\includegraphics[width=1.25\linewidth]{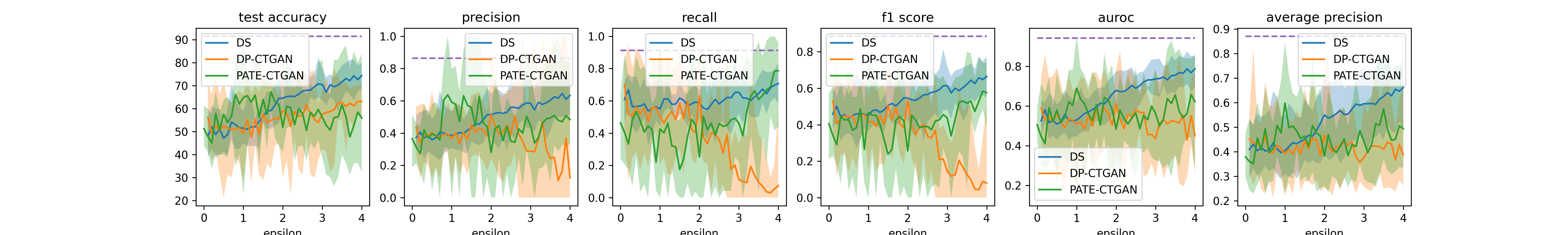}}
    \caption{Social Network with KNN}
    \label{sn knn}
\end{figure}

\begin{figure}[ht]
    \centering
    \makebox[\textwidth][c]{\includegraphics[width=1.25\linewidth]{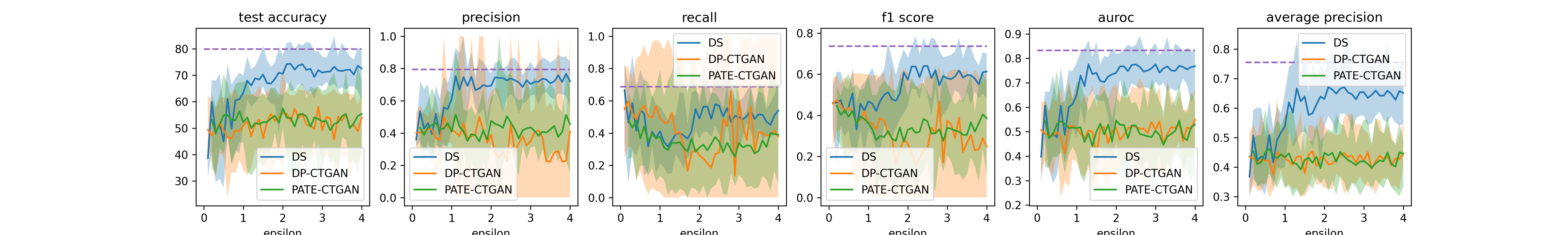}}
    \caption{Titanic with KNN}
    \label{t knn}
\end{figure}

\textbf{KNN}
Despite of results on Banknote \ref{banknote knn}, here are results on KNN classifier on Iris \ref{iris knn}, Social Network \ref{sn knn} and Titanic \ref{t knn}. PATE-CTGAN exceeds DataSynthesizer with large $\varepsilon$ on Iris \ref{iris knn}. Different with previous classifier, DataSynthesizer show a slight recovery trend on Recall. Results on Social Network \ref{sn knn} and Titanic \ref{t knn} are similar with previous results.

\begin{figure}[ht]
    \centering
    \makebox[\textwidth][c]{\includegraphics[width=1.25\linewidth]{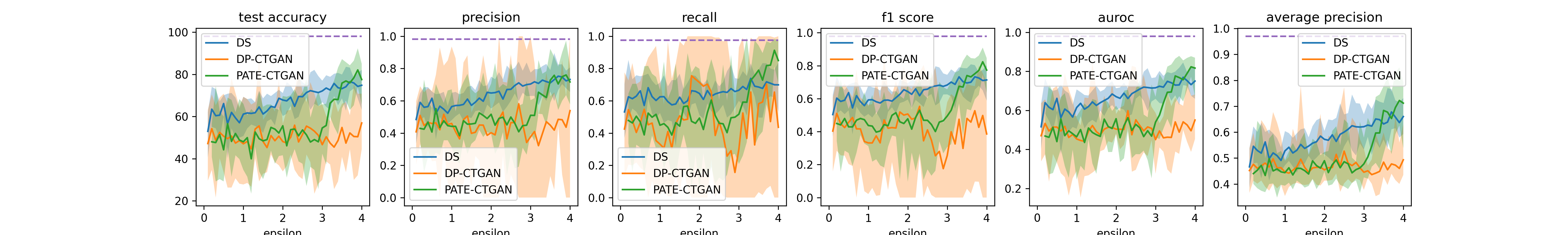}}
    \caption{Banknote with Decision Tree}
    \label{banknote dt}
\end{figure}

\begin{figure}[ht]
    \centering
    \makebox[\textwidth][c]{\includegraphics[width=\linewidth]{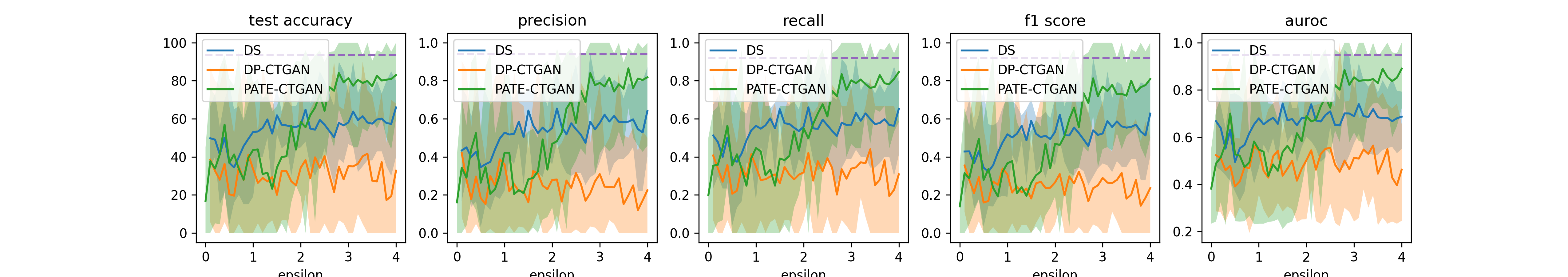}}
    \caption{Iris with Decision Tree}
    \label{iris dt}
\end{figure}

\begin{figure}[ht]
    \centering
    \makebox[\textwidth][c]{\includegraphics[width=1.25\linewidth]{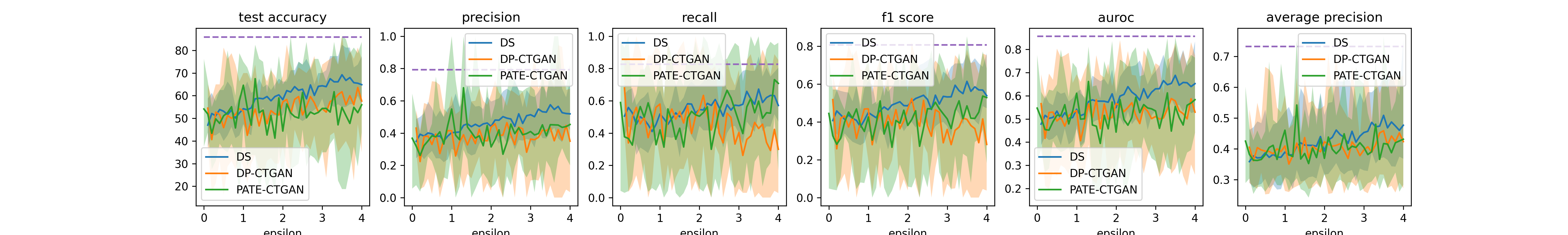}}
    \caption{Social Network with Decision Tree}
    \label{sn dt}
\end{figure}

\begin{figure}[ht]
    \centering
    \makebox[\textwidth][c]{\includegraphics[width=1.25\linewidth]{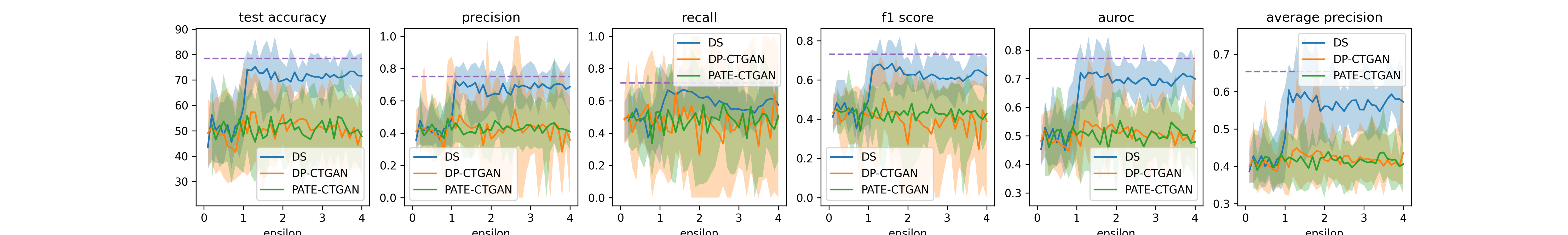}}
    \caption{Titanic with Decision Tree}
    \label{t dt}
\end{figure}

\textbf{Decision Tree}
DataSynthesizer is still the best one with Decision Tree on Banknote \ref{banknote dt}, Social Network \ref{sn dt} and Titanic \ref{t dt}. PATE-CTGAN also has good performance on Iris \ref{iris dt} with large $\varepsilon$. 

\begin{figure}[ht]
    \centering
    \makebox[\textwidth][c]{\includegraphics[width=1.25\linewidth]{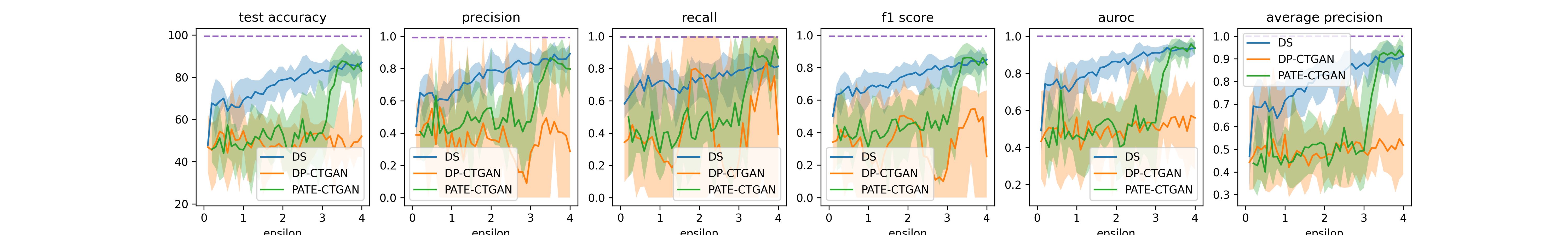}}
    \caption{Banknote with Random Forest}
    \label{banknote rf}
\end{figure}

\begin{figure}[ht]
    \centering
    \makebox[\textwidth][c]{\includegraphics[width=\linewidth]{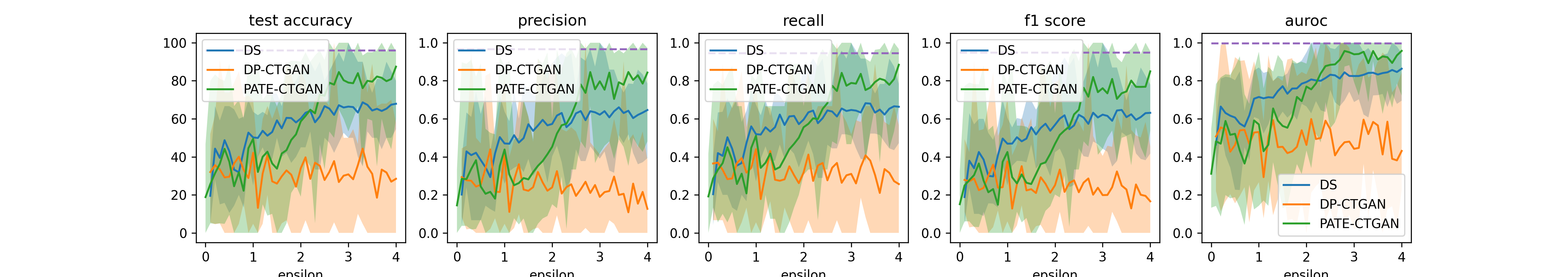}}
    \caption{Iris with Random Forest}
    \label{iris rf}
\end{figure}

\begin{figure}[ht]
    \centering
    \makebox[\textwidth][c]{\includegraphics[width=1.25\linewidth]{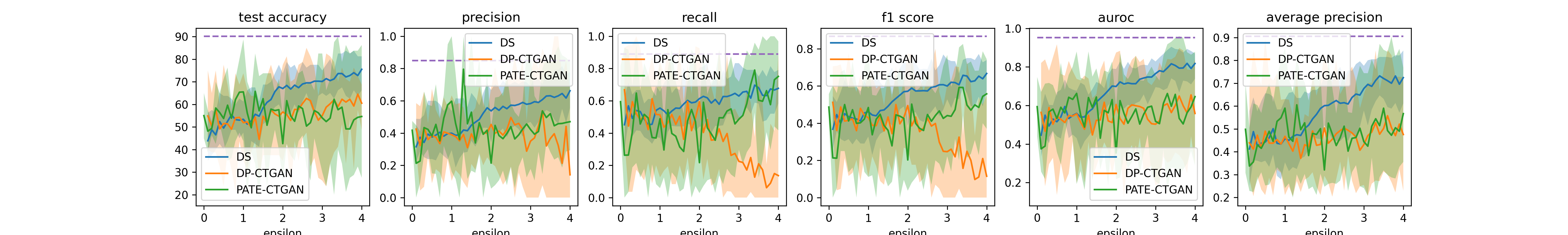}}
    \caption{Social Network with Random Forest}
    \label{sn rf}
\end{figure}

\begin{figure}[ht]
    \centering
    \makebox[\textwidth][c]{\includegraphics[width=1.25\linewidth]{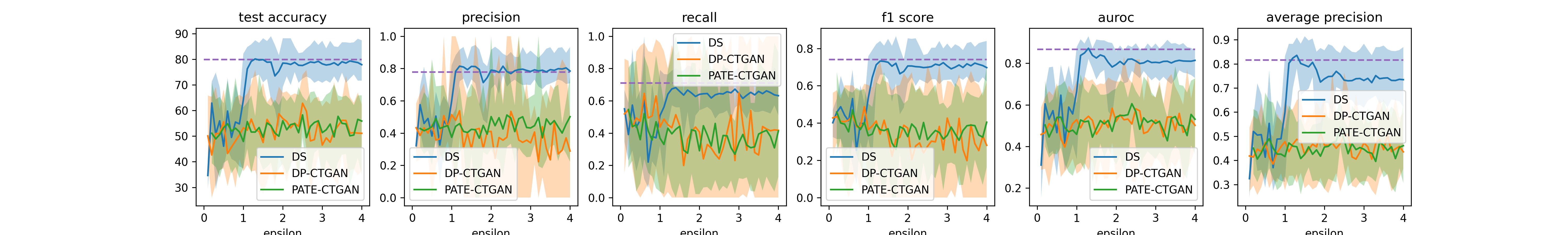}}
    \caption{Titanic with Random Forest}
    \label{t rf}
\end{figure}

\textbf{Random Forest}
DataSynthesizer has the utility recovery phenonmenon on Recall metric with this Random Forest method, see figures for Banknote \ref{banknote rf}, Iris \ref{iris rf}, Social Network \ref{sn rf} and Titanic \ref{t rf}. PATE-CTGAN defeats DataSynthesizer on Iris with $\varepsilon > 2$.

\begin{figure}[ht]
    \centering
    \makebox[\textwidth][c]{\includegraphics[width=1.25\linewidth]{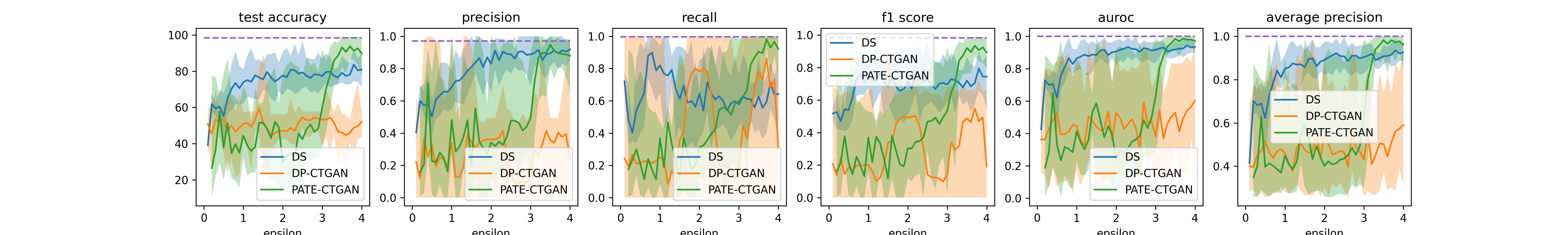}}
    \caption{Banknote with Logistics Regression}
    \label{banknote lr}
\end{figure}

\begin{figure}[ht]
    \centering
    \makebox[\textwidth][c]{\includegraphics[width=\linewidth]{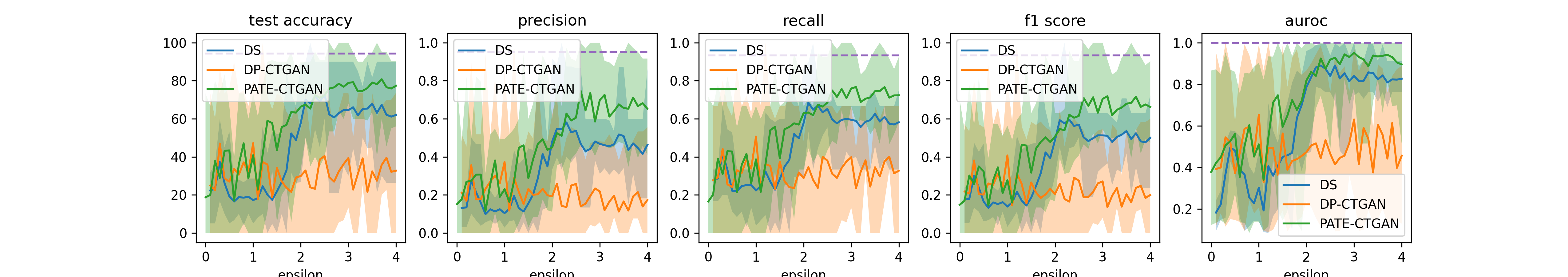}}
    \caption{Iris with Logistics Regression}
    \label{iris lr}
\end{figure}

\begin{figure}[ht]
    \centering
    \makebox[\textwidth][c]{\includegraphics[width=1.25\linewidth]{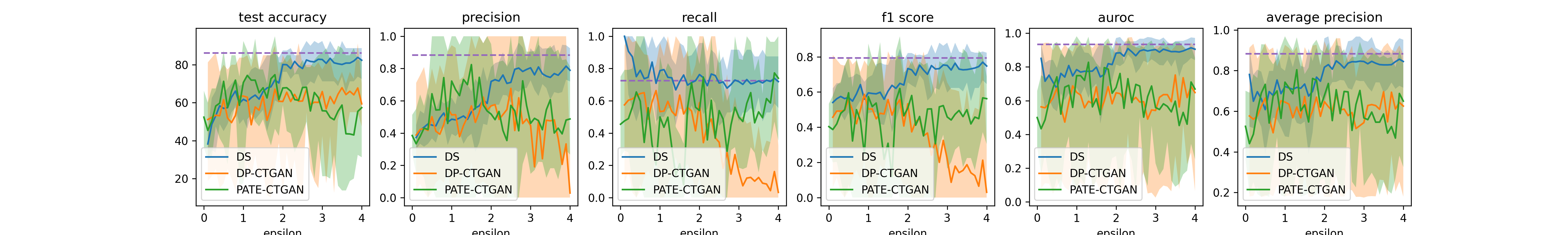}}
    \caption{Social Network with Logistics Regression}
    \label{sn lr}
\end{figure}

\begin{figure}[ht]
    \centering
    \makebox[\textwidth][c]{\includegraphics[width=1.25\linewidth]{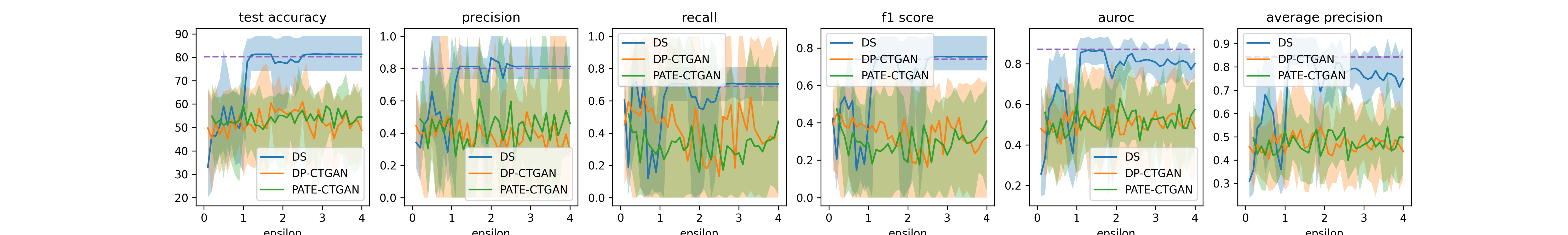}}
    \caption{Titanic with Logistics Regression}
    \label{t lr}
\end{figure}

\textbf{Logistic Regression}
Different result with logistic regression is that DataSynthesizer doesn't has the recovery trend on Recall. But it has better Recall score than Original training data on Social Network \ref{sn lr} and Titanic \ref{t lr}. Results on Banknote \ref{banknote lr} and Iris \ref{iris lr} are similar with other methods.

\clearpage

\section{More Results for Utility Recovery Incapability phenomenon across different source real dataset}\label{appendix:dataset}
Since there is no Average Precision on Iris dataset, we don't take account for Average Precision in this section.

\begin{figure}[ht]
    \centering
    \includegraphics[width=\linewidth]{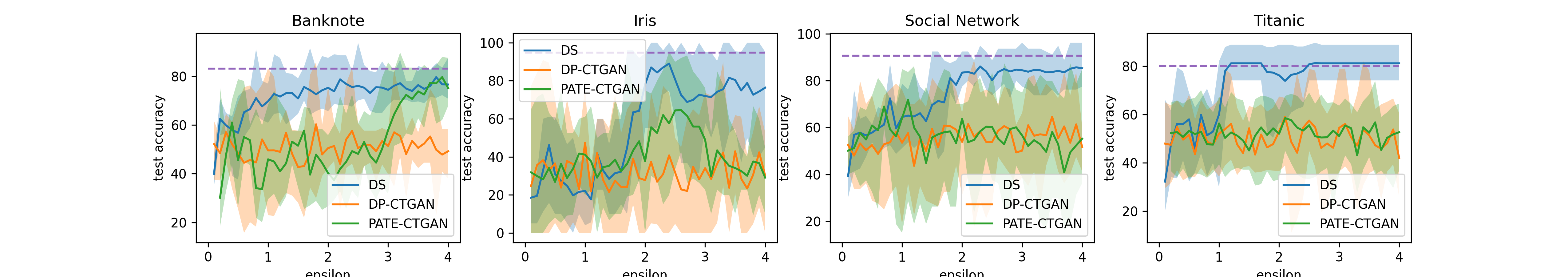}
    \caption{Accuracy with Na\"ive Bayes}
    \label{anb}
\end{figure}

\begin{figure}[ht]
    \centering
    \includegraphics[width=\linewidth]{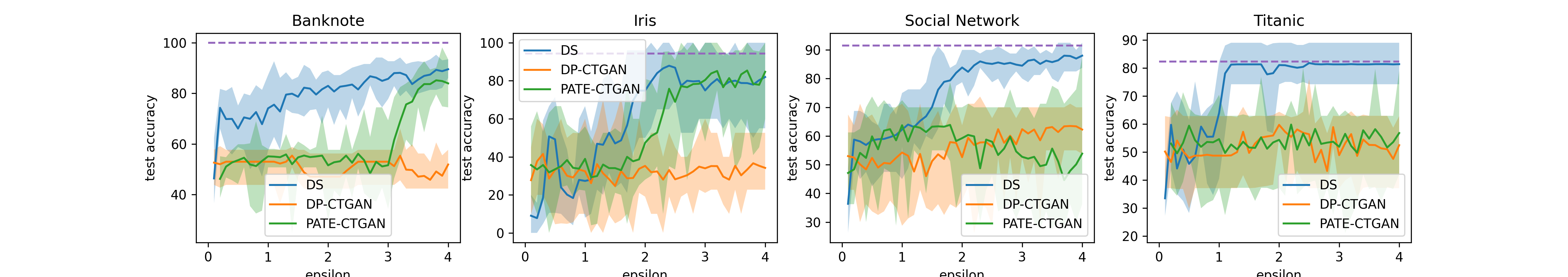}
    \caption{Accuracy with SVM}
    \label{asvm}
\end{figure}

\begin{figure}[ht]
    \centering
    \includegraphics[width=\linewidth]{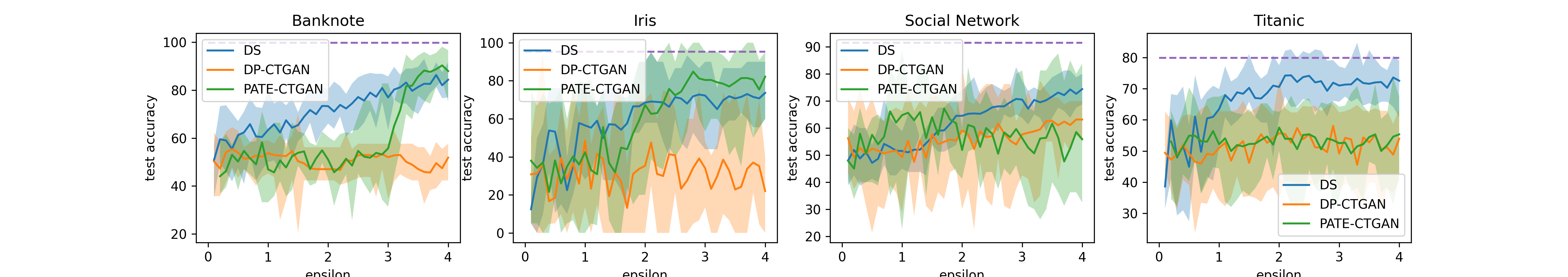}
    \caption{Accuracy with KNN}
    \label{aknn}
\end{figure}

\begin{figure}[ht]
    \centering
    \includegraphics[width=\linewidth]{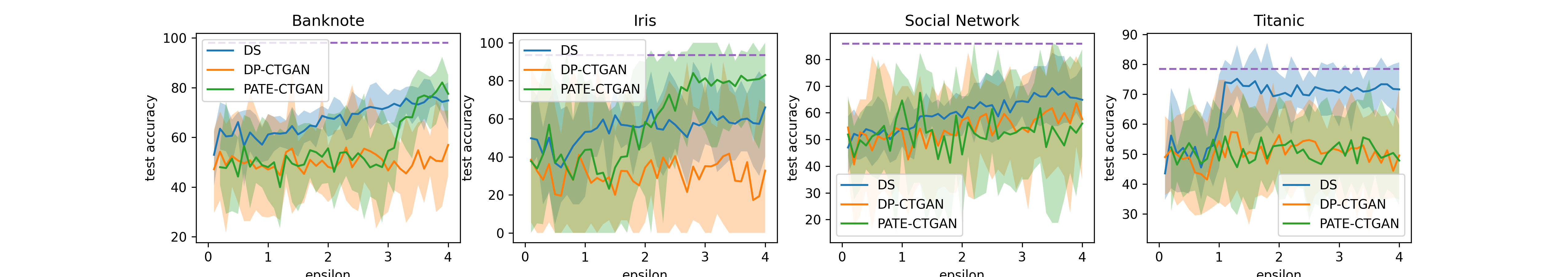}
    \caption{Accuracy with Decision tree}
    \label{adt}
\end{figure}

\begin{figure}[ht]
    \centering
    \includegraphics[width=\linewidth]{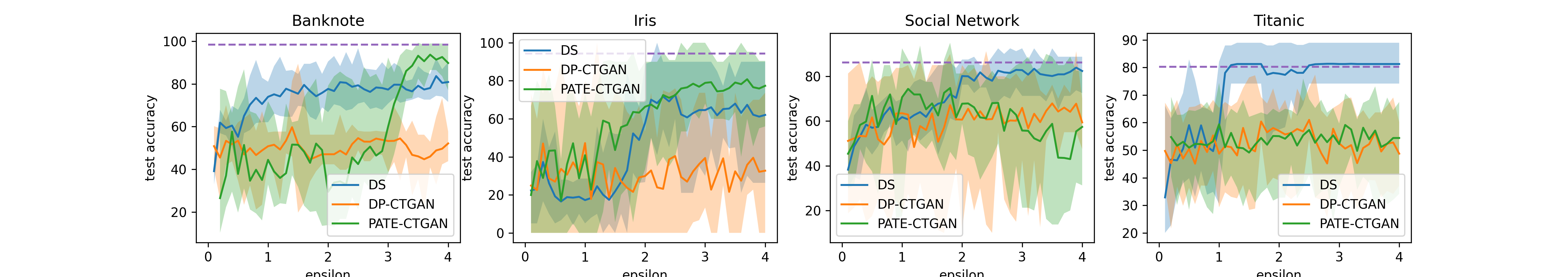}
    \caption{Accuracy with Logistic Regression}
    \label{alr}
\end{figure}

\textbf{Accuracy}
For Accuracy, DataSynthesizer has better performance with Na\"ive Bayes \ref{anb}, SVM \ref{asvm}, and Logistic Regression \ref{alr} for all four datasets. PATE-CTGAN has good performance on Iris with KNN \ref{aknn}, Decision Tree \ref{adt}. Results with Random Forest \ref{arf} has been shown before. DP-CTGAN always does badly on this work.

\begin{figure}[ht]
    \centering
    \includegraphics[width=\linewidth]{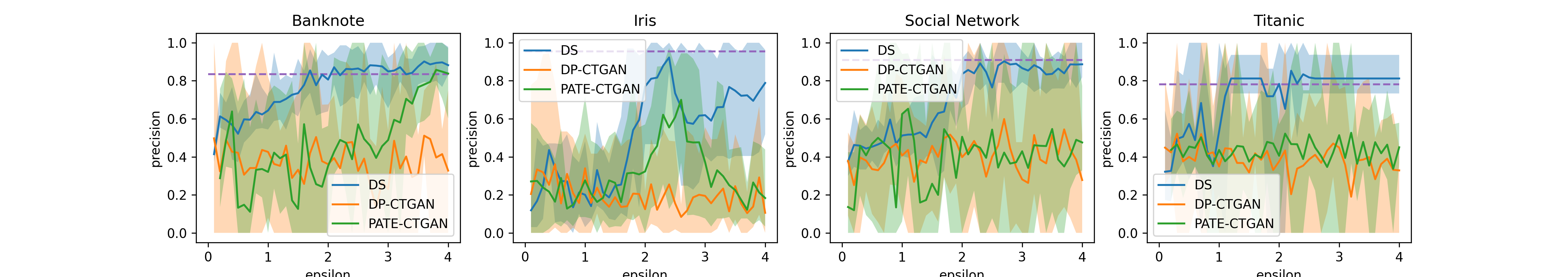}
    \caption{Precision with Na\"ive Bayes}
    \label{pnb}
\end{figure}

\begin{figure}[ht]
    \centering
    \includegraphics[width=\linewidth]{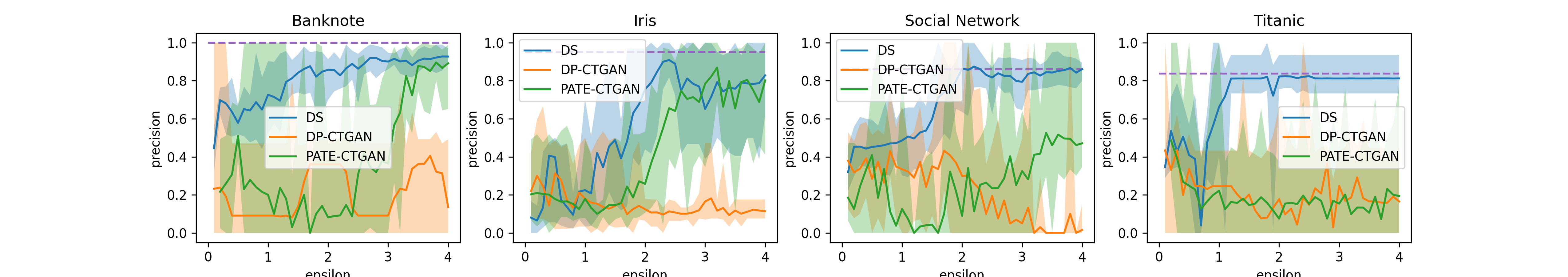}
    \caption{Precision with SVM}
    \label{psvm}
\end{figure}

\begin{figure}[ht]
    \centering
    \includegraphics[width=\linewidth]{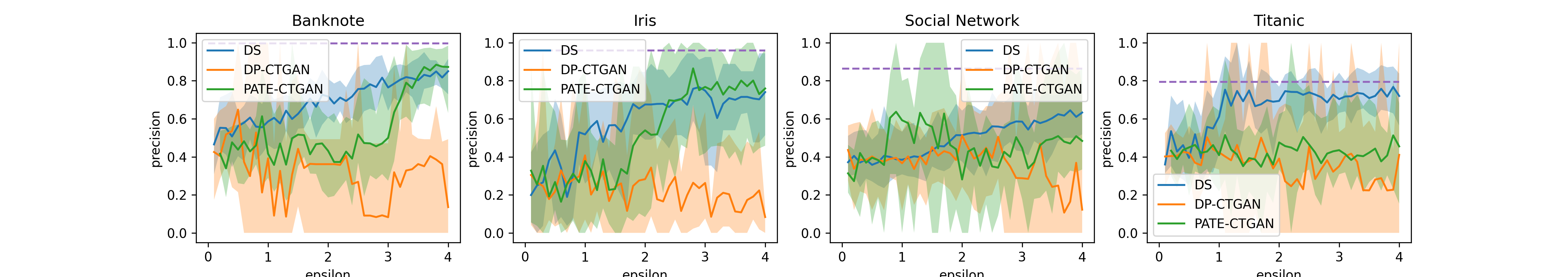}
    \caption{Precision with KNN}
    \label{pknn}
\end{figure}

\begin{figure}[ht]
    \centering
    \includegraphics[width=\linewidth]{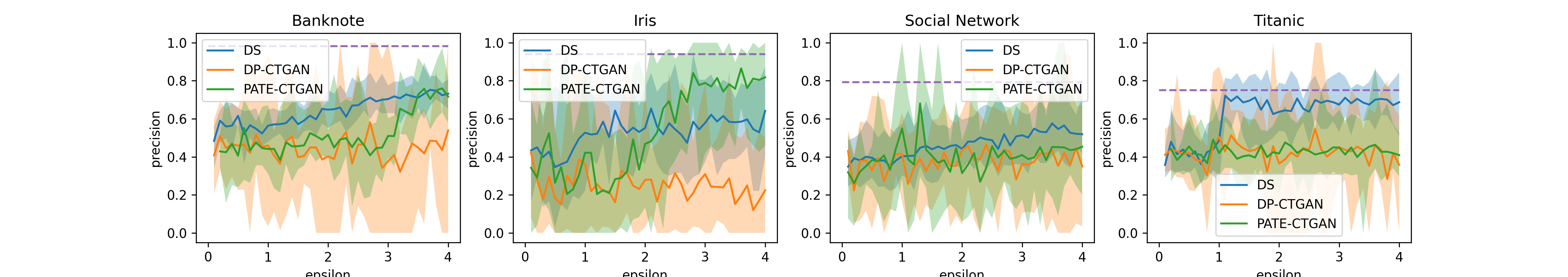}
    \caption{Precision with Decision tree}
    \label{pdt}
\end{figure}

\begin{figure}[ht]
    \centering
    \includegraphics[width=\linewidth]{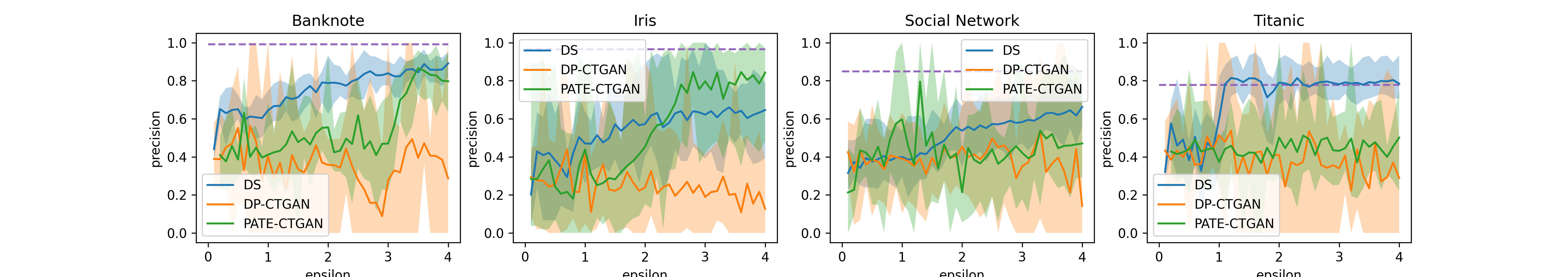}
    \caption{Precision with Random Forest}
    \label{prf}
\end{figure}

\begin{figure}[ht]
    \centering
    \includegraphics[width=\linewidth]{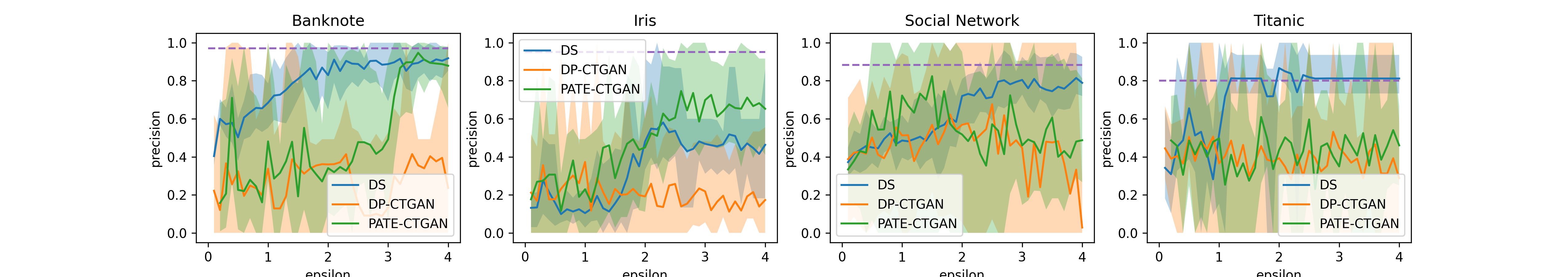}
    \caption{Precision with Logistic Regression}
    \label{plr}
\end{figure}

\textbf{Precision}
DataSynthesizer is still roughly consistent on four datasets, while PATE-CTGAN doesn't work well on Social Network and Titanic. See figures below for Na\"ive Bayes \ref{pnb}, SVM \ref{psvm}, KNN \ref{pknn}, Decision tree \ref{pdt}, Random Forest \ref{prf} and Logistic Regression \ref{plr}.

\begin{figure}[ht]
    \centering
    \includegraphics[width=\linewidth]{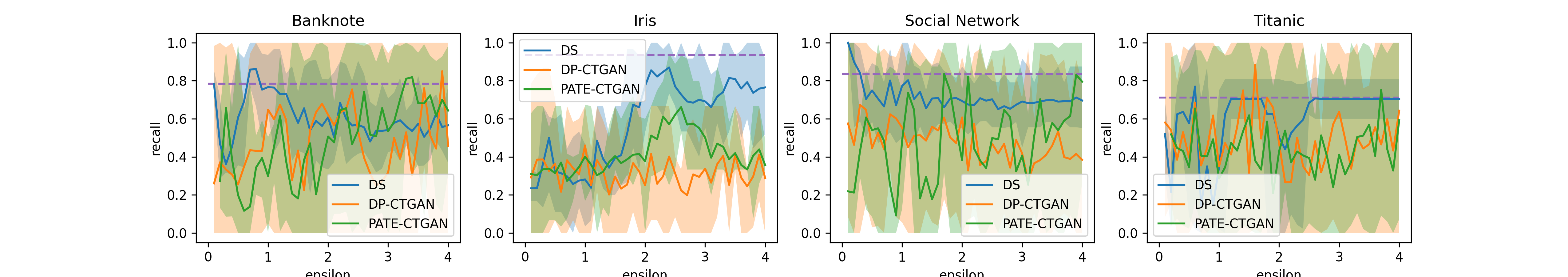}
    \caption{Recall with Na\"ive Bayes}
    \label{rnb}
\end{figure}

\begin{figure}[ht]
    \centering
    \includegraphics[width=\linewidth]{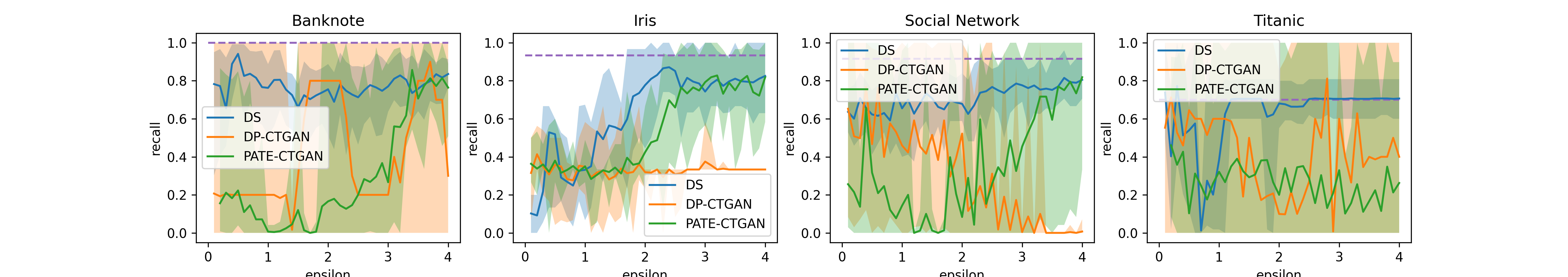}
    \caption{Recall with SVM}
    \label{rsvm}
\end{figure}

\begin{figure}[ht]
    \centering
    \includegraphics[width=\linewidth]{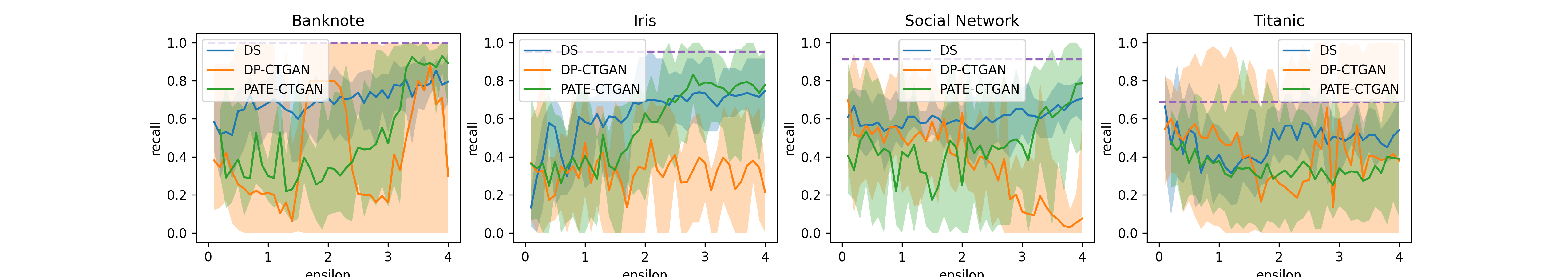}
    \caption{Recall with KNN}
    \label{rknn}
\end{figure}

\begin{figure}[ht]
    \centering
    \includegraphics[width=\linewidth]{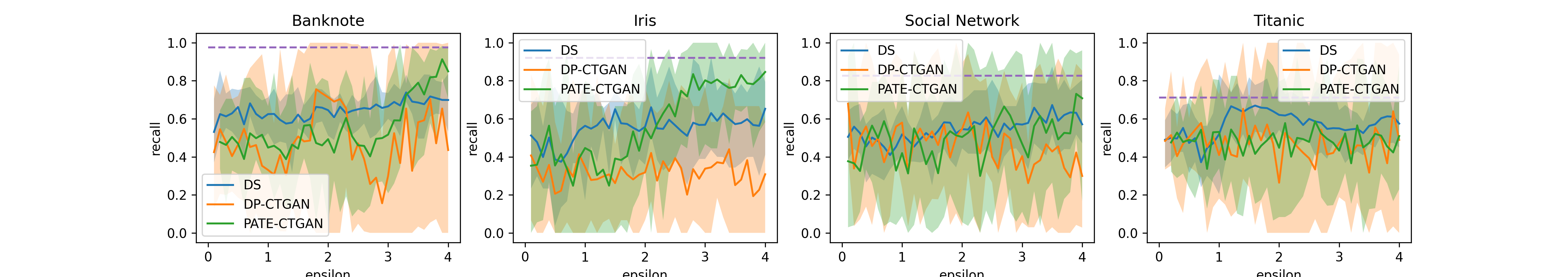}
    \caption{Recall with Decision tree}
    \label{rdt}
\end{figure}

\begin{figure}[ht]
    \centering
    \includegraphics[width=\linewidth]{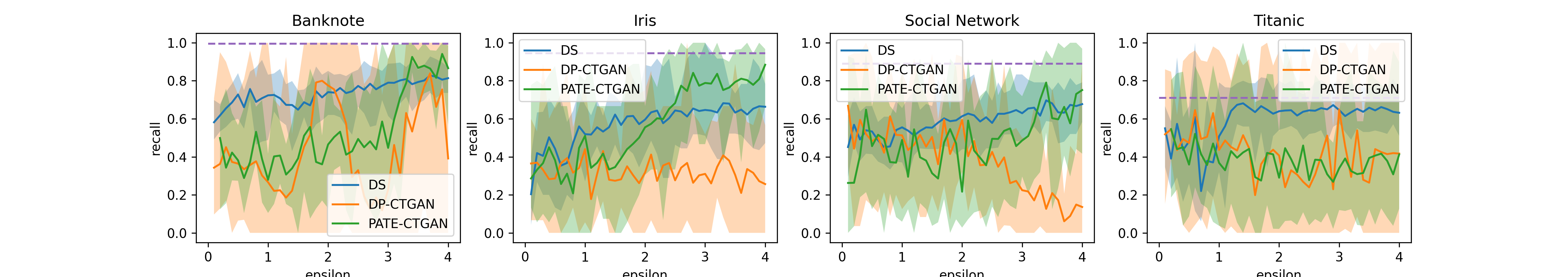}
    \caption{Recall with Random Forest}
    \label{rrf}
\end{figure}

\begin{figure}[ht]
    \centering
    \includegraphics[width=\linewidth]{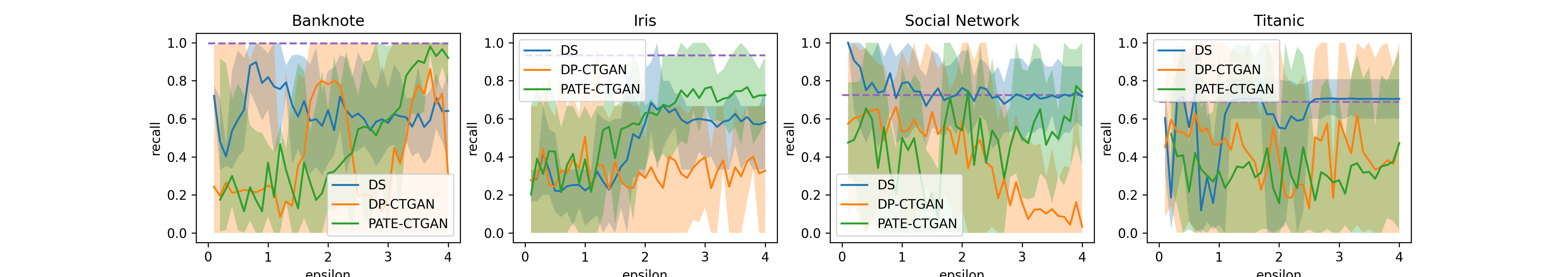}
    \caption{Recall with Logistic Regression}
    \label{rlr}
\end{figure}

\textbf{Recall}
The Utility recovery phenomenon is not obvious on Recall metric. See figures below for Na\"ive Bayes \ref{rnb}, SVM \ref{rsvm}, KNN \ref{rknn}, Decision tree \ref{rdt}, Random Forest \ref{rrf} and Logistic Regression \ref{rlr}.

\begin{figure}[ht]
    \centering
    \includegraphics[width=\linewidth]{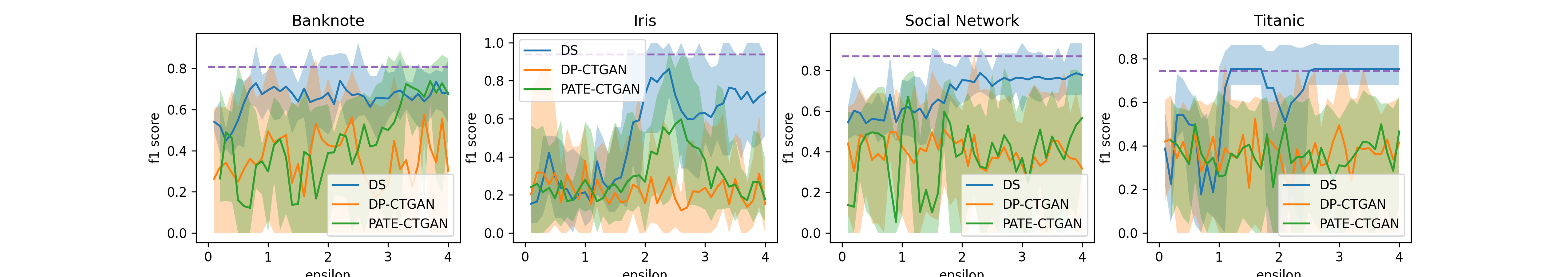}
    \caption{F1 Score with Na\"ive Bayes}
    \label{fnb}
\end{figure}

\begin{figure}[ht]
    \centering
    \includegraphics[width=\linewidth]{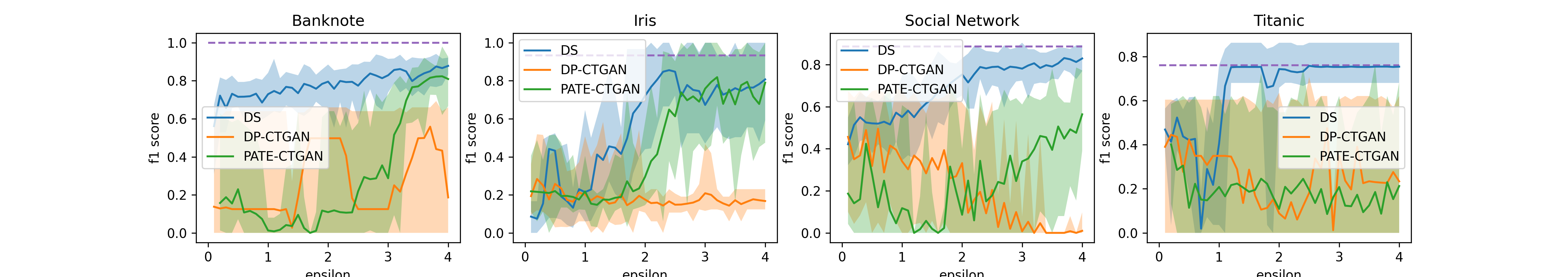}
    \caption{F1 Score with SVM}
    \label{fsvm}
\end{figure}

\begin{figure}[ht]
    \centering
    \includegraphics[width=\linewidth]{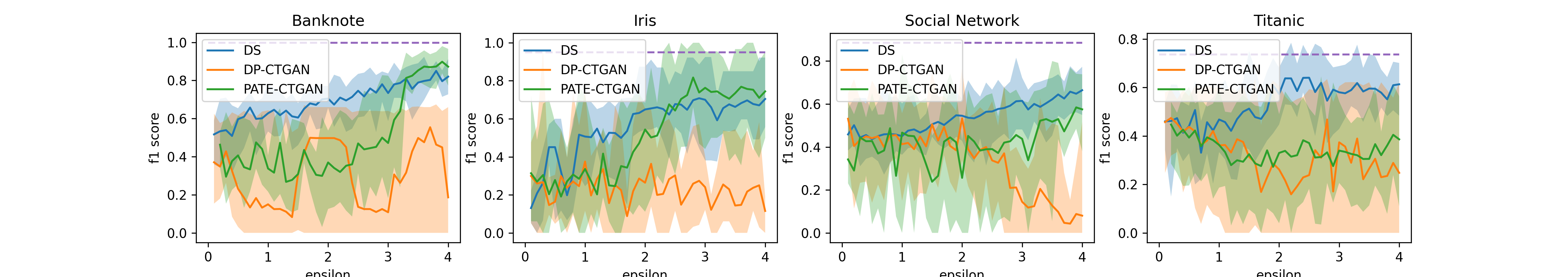}
    \caption{F1 Score with KNN}
    \label{fknn}
\end{figure}

\begin{figure}[ht]
    \centering
    \includegraphics[width=\linewidth]{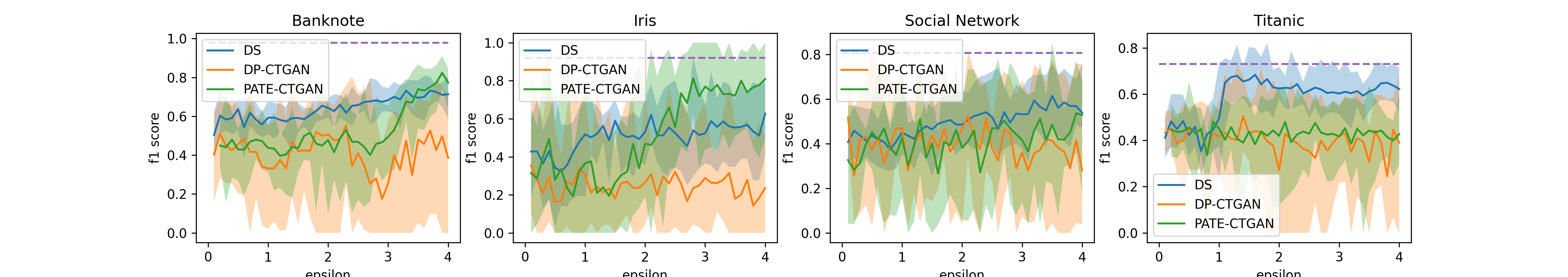}
    \caption{F1 Score with Decision tree}
    \label{fdt}
\end{figure}

\begin{figure}[ht]
    \centering
    \includegraphics[width=\linewidth]{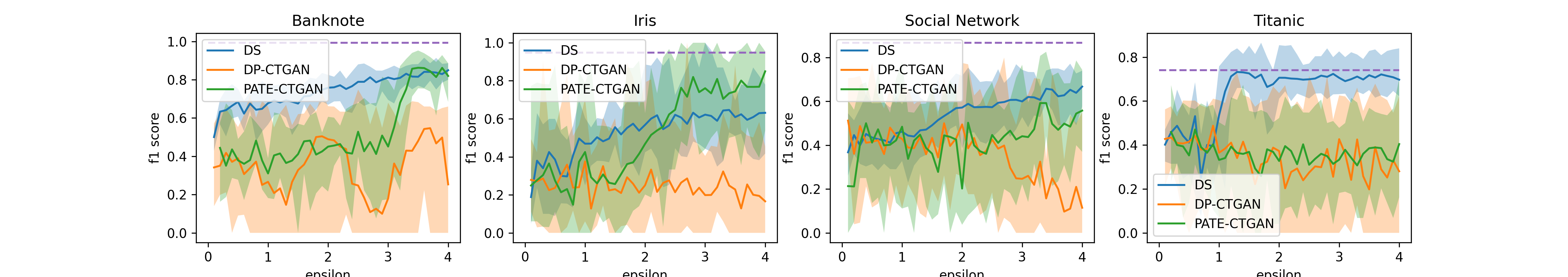}
    \caption{F1 Score with Random Forest}
    \label{frf}
\end{figure}

\begin{figure}[ht]
    \centering
    \includegraphics[width=\linewidth]{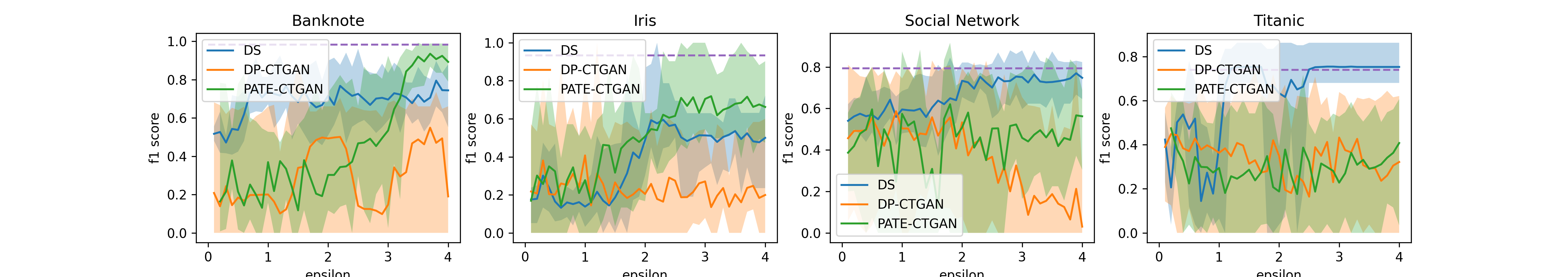}
    \caption{F1 Score with Logistic Regression}
    \label{flr}
\end{figure}

\textbf{F1 Score}
Results for F1 Score are similar, see here for Na\"ive Bayes \ref{fnb}, SVM \ref{fsvm}, KNN \ref{fknn}, Decision tree \ref{fdt}, Random Forest \ref{frf} and Logistic Regression \ref{flr}.

\begin{figure}[ht]
    \centering
    \includegraphics[width=\linewidth]{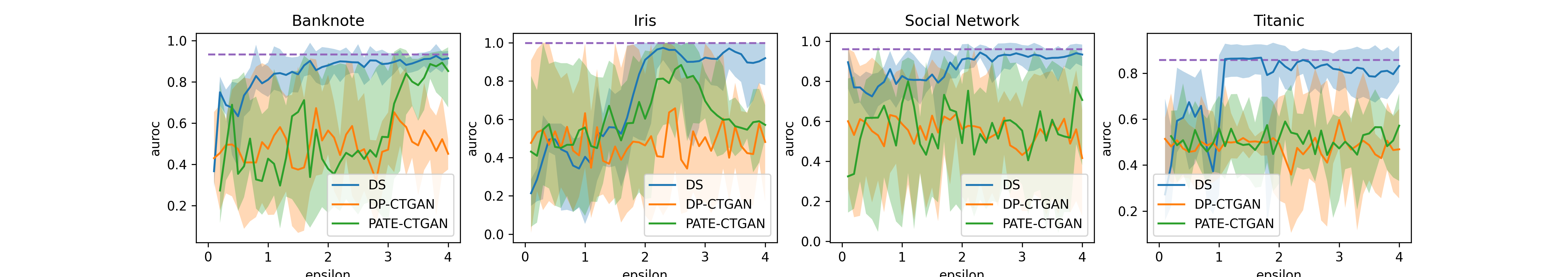}
    \caption{AUROC with Na\"ive Bayes}
    \label{aurocnb}
\end{figure}

\begin{figure}[ht]
    \centering
    \includegraphics[width=\linewidth]{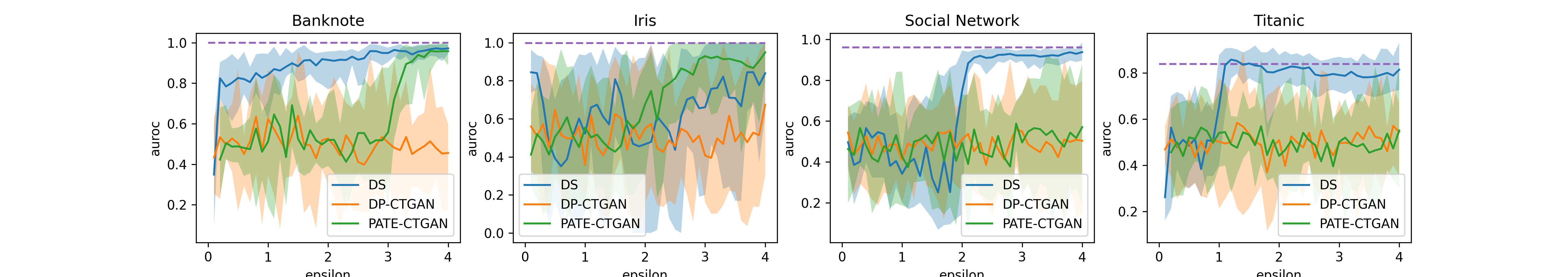}
    \caption{AUROC with SVM}
    \label{aurocsvm}
\end{figure}

\begin{figure}[ht]
    \centering
    \includegraphics[width=\linewidth]{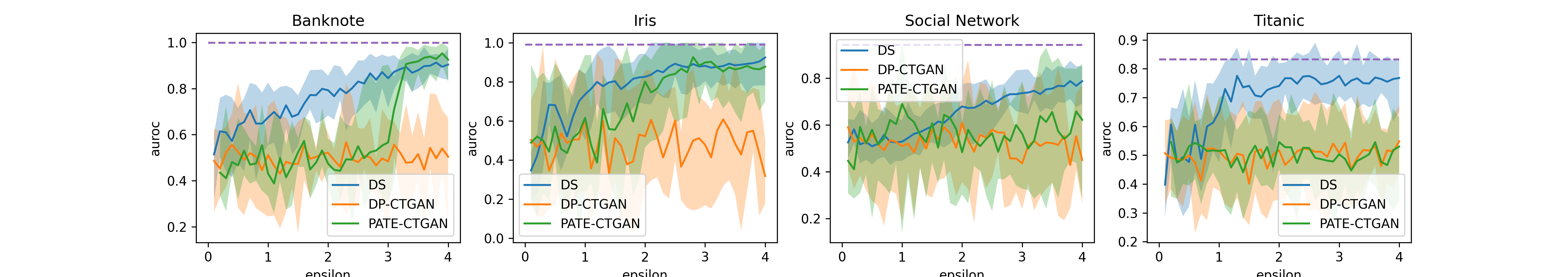}
    \caption{AUROC with KNN}
    \label{aurocknn}
\end{figure}

\begin{figure}[ht]
    \centering
    \includegraphics[width=\linewidth]{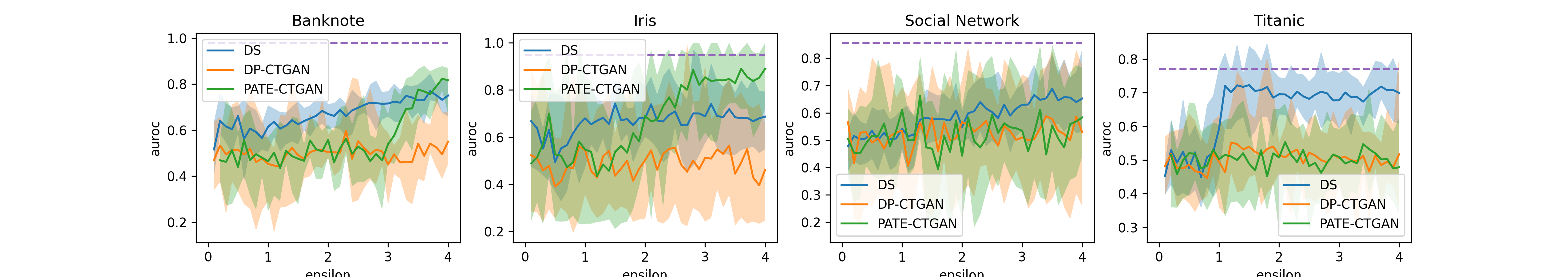}
    \caption{AUROC with Decision tree}
    \label{aurocdt}
\end{figure}

\begin{figure}[ht]
    \centering
    \includegraphics[width=\linewidth]{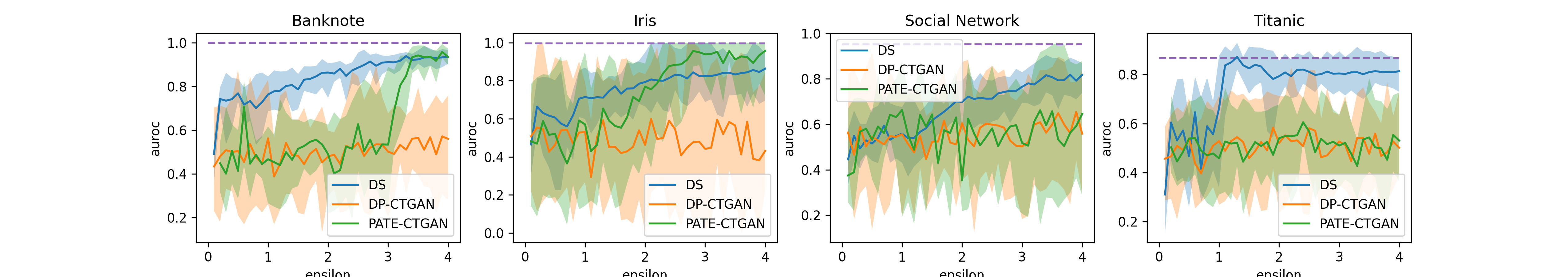}
    \caption{AUROC with Random Forest}
    \label{aurocrf}
\end{figure}

\begin{figure}[ht]
    \centering
    \includegraphics[width=\linewidth]{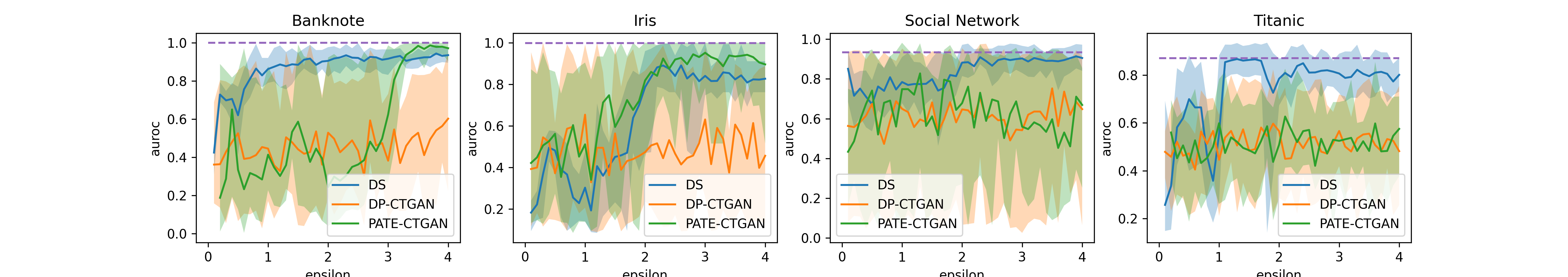}
    \caption{AUROC with Logistic Regression}
    \label{auroclr}
\end{figure}

\textbf{AUROC}
As mentioned before, PATE-CTGAN can only work well on Banknote and Iris. It is heavily influenced by the dataset. See figures for Na\"ive Bayes \ref{aurocnb}, SVM \ref{aurocsvm}, KNN \ref{aurocknn}, Decision tree \ref{aurocdt}, Random Forest \ref{aurocrf} and Logistic Regression \ref{auroclr}.

\clearpage